\documentclass[twoside]{article}

\usepackage[accepted]{aistats2026}

\usepackage{microtype}
\usepackage{graphicx}
\usepackage{subfigure}
\usepackage{booktabs}

\usepackage{hyperref}

\usepackage{amsmath}
\usepackage{amssymb}
\usepackage{mathtools}
\usepackage{amsthm}
\usepackage[capitalize,noabbrev]{cleveref}

\theoremstyle{plain}
\newtheorem{theorem}{Theorem}[section]
\newtheorem{proposition}[theorem]{Proposition}

\theoremstyle{definition}

\newtheorem{assumption}[theorem]{Assumption}
\theoremstyle{remark}

\usepackage{amsfonts}
\usepackage{nicefrac}
\usepackage{natbib}
\usepackage{doi}
\usepackage{bbm}
\usepackage{enumitem}
\usepackage{appendix}
\usepackage{tikz}
\usetikzlibrary{shapes, arrows.meta}
\newcommand{\ind}{\mathbbm{1}}
\newcommand{\x}{{\boldsymbol{x}}}
\newcommand{\X}{{\boldsymbol{X}}}
\newcommand{\w}{{\boldsymbol{w}}}

\begin{document}
\runningauthor{Arbour, Parikh, Niknam, Stuart, Rudolph, Feller}

\twocolumn[
\aistatstitle{Regularizing Extrapolation in Causal Inference}

\aistatsauthor{David Arbour\footnotemark[1] \And Harsh Parikh\footnotemark[1] \And  Bijan Niknam}

\aistatsaddress{Adobe Research \And  Yale University \And Johns Hopkins University}

\aistatsauthor{Elizabeth Stuart \And Kara Rudolph \And Avi Feller}
\aistatsaddress{Johns Hopkins University \And Columbia University \And University of California, Berkeley}
]
\footnotetext[1]{co-first authors; alphabetical order}

\begin{abstract}
Many causal inference and machine learning estimators are linear smoothers, where the prediction is a weighted average of training outcomes. Whether weights are constrained to be non-negative creates a key tradeoff: non-negative weights (e.g., inverse propensity weighting, random forests) limit extrapolation but can worsen covariate imbalance, while unconstrained weights (e.g., OLS, kernel ridge regression) improve balance but increase dependence on parametric assumptions. We propose a unified framework that directly penalizes extrapolation via a soft constraint on negative weights, replacing the standard hard non-negativity restriction. We derive a worst-case error bound and introduce a novel ``bias-bias-variance'' tradeoff among distributional imbalance, model misspecification, and estimator variance; this tradeoff is especially pronounced in high dimensions with poor positivity. We develop a convex optimization procedure that regularizes this bound and outline how to use the extrapolation penalty as a sensitivity analysis for parametric assumptions. We demonstrate our approach on synthetic data and a real-world application generalizing randomized trial estimates to a target population.
\end{abstract}

\section{INTRODUCTION}
A core challenge in observational causal inference and domain adaptation is to adjust data distributions so that features are comparable across distinct groups, such as control and treated arms or source and target populations \citep{imbens2015causal, farahani2021brief}. Weighting estimators and linear smoothers, in which the prediction is a weighted average of training outcomes, are widely used for such adjustment; examples include implicit weighting estimators like ordinary least squares (OLS) and random forests and explicit weighting approaches like inverse propensity score weighting \citep{li2013propensity} and importance sampling \citep{thomas2017importance}.

An important divide among weighting estimators is whether weights are constrained to be non-negative, such as in traditional IPW, matching \citep{stuart2010}, the synthetic control method \citep{abadie2010synthetic}, and stable balancing weights \citep{zubizarreta2015stable, ben2021balancing}, as well as in the weighting component of popular doubly robust estimators like double machine learning \citep{chernozhukov2018double}.
This constraint limits extrapolation and dependence on parametric modeling assumptions, but typically at the cost of worse feature imbalance between re-weighted groups. This imbalance is especially pronounced in high-dimensional settings, when the curse of dimensionality means that positivity is less likely to hold, leading to further bias \citep{damour2021overlap}.
By contrast, linear smoothers like OLS and kernel ridge regression allow for arbitrarily negative weights \citep{robins2007comment}, which can improve feature imbalance but at the cost of greater model dependence and higher estimator variance.
Finally, augmented estimators that combine outcome modeling with explicit weighting strategies can be viewed as performing controlled extrapolation, balancing model dependence against feature imbalance. Pure weighting and pure outcome modeling thus represent the two extremes of no versus uncontrolled extrapolation.

In this paper, we leverage this geometric perspective to establish a general framework for systematically controlling extrapolation.
In particular, we propose a unified approach that directly penalizes the level of extrapolation, replacing the current practice of a hard non-negativity constraint with a soft constraint and corresponding hyperparameter.
Unlike prior research on extrapolation in machine learning that emphasizes predictions beyond observed covariate support, we conceptualize extrapolation through unit weights, a particularly natural framework for handling high-dimensional covariates \citep{ben2021augmented}.
Specifically, our contributions are:
\begin{itemize}[leftmargin=*]
\item \textit{Bias-bias-variance tradeoff.} We propose a framework quantifying a ``bias-bias-variance" tradeoff, decomposing error into bias from distributional imbalance, bias from outcome model misspecification, and estimator variance. This captures key tradeoffs encountered in common causal inference and distribution shift scenarios.

\item \textit{Error bound and constrained optimization.} We derive an error bound based on worst-case H{\"o}lder continuity deviations from linearity. We present an optimization approach to minimize this bound, explicitly controlling tradeoffs between biases. We characterize the finite-sample variance through our error bounds.

\item \textit{Sensitivity analysis framework.} We introduce a sensitivity analysis methodology integrated into our optimization framework, enabling systematic evaluation of distributional imbalance and outcome model misspecification impacts. We illustrate this using synthetic data and a practical application involving the transportation of causal estimates to a novel target population.
\end{itemize}

\subsection{Related work}
Extrapolation and generalization are core topics in causal inference and machine learning. Recent surveys by \citet{degtiar2023review} and \citet{johansson2022generalization} provide comprehensive overviews on generalizability and transportability methods.

\paragraph{Extrapolation and the synthetic control method.} Extrapolating far from the support of the data is a longstanding concern in statistics and the social sciences especially; see \citet{king2006dangers} for a seminal discussion of possible dangers of unchecked extrapolation. Methods that limit extrapolation are common; the synthetic control method \citep{abadie2010synthetic} is a particularly prominent example. \citet{doudchenko2016balancing} discuss the non-negativity constraint in this context, and explore possible regularization. Most relevant to our approach, \citet{ben2021augmented} developed the augmented synthetic control method, which combines outcome modeling with constrained weights to reduce bias while controlling extrapolation.

\paragraph{Extrapolation in machine learning.}
Within machine learning, there has been substantial recent progress on approaches for addressing extrapolation. \citet{shen2024engression} introduced engression, a framework that views extrapolation through the lens of distributional regression, enabling principled uncertainty quantification outside the training distribution. \citet{kong2024towards} developed a causal lens for understanding extrapolation, establishing theoretical connections between causal structure and extrapolation. \citet{netanyahu2023learning} proposed a transductive approach for learning to extrapolate, leveraging unlabeled test points to guide the extrapolation process. \citet{dong2022first} provided foundational analysis toward understanding the extrapolation of nonlinear models to unseen domains, establishing bounds on extrapolation error. Finally, \citet{pfister2024extrapolation} developed extrapolation-aware nonparametric statistical inference methods, with formal guarantees on validity beyond the support of training data.

Unlike this recent literature, we approach extrapolation from a weighting perspective, which offers particular advantages in high-dimensional settings. Rather than focusing on predictions outside the covariate support, we frame extrapolation in terms of the properties of unit weights, providing a natural parameterization for high-dimensional settings \citep{ben2021augmented}. This perspective allows us to directly quantify and regularize the degree of extrapolation without relying on complex directional derivatives or high-dimensional density estimation.

\paragraph{Positivity violations and shifting the target.} Our discussion is closely related to the literature on positivity violations in causal inference.
\citet{crump2006moving}, \citet{li2018balancing}, and \citet{parikh2024we} all proposed to avoid issues due to positivity violations by shifting the estimand to regions with greater overlap. By contrast, our approach directly incorporates the severity of positivity violations into the weight estimation process.

\paragraph{Weighting representations.} A growing literature highlights the connections between various causal estimators through their weighting representations \citep{chattopadhyay2023implied}. \citet{knaus2024treatment} provided a unified framework for viewing treatment effect estimators as weighted outcomes. \citet{bruns2023augmented} showed that augmented balancing weights can be interpreted as a form of linear regression. \citet{lin2022regression} examined regression-adjusted imputation estimators through their weighting properties. Our framework builds on these insights by explicitly parameterizing the degree of extrapolation through weight regularization, providing a continuum of estimators that navigate the bias-variance tradeoff.

\section{PRELIMINARIES}

\subsection{Setup and notation}

To ease exposition, we focus on the causal inference problem of estimating the missing control potential outcome for the Average Treatment Effect on the Treated (ATT). As we note below, however, these results hold for general linear estimands as well as for domain adaptation in ML \citep{johansson2022generalization}. 

For each unit $i \in [n]$, we observe the tuple $(X_i, Y_i, Z_i)$, with covariates $X_i \in \mathcal{X}$, outcome $Y_i \in \mathbb{R}$, and binary treatment $Z_i \in \{0,1\}$. 
Invoking SUTVA, let $Y_i(0)$ and $Y_i(1)$ denote the control and treated potential outcomes, respectively, for unit $i$.
Our estimand of interest is the ATT, $\mathbb{E}[Y_i(1) - Y_i(0) \mid Z_i = 1]$; we discuss generalizations below. Since we observe $Y(1)$ for the treated group, the key challenge is to estimate the missing control potential outcome mean, $\mathbb{E}[Y_i(0) \mid Z_i = 1]$. Finally, define the density ratio $dQ/dP(x)$, where $Q$ and $P$ denote the populations of units assigned to treatment and control, respectively. 
We need the following key assumptions for nonparametric identification:
\begin{enumerate}[label = A.\arabic*., leftmargin=*]
    \item (Exchangeability) $\mathbb{E}[Y(0) \mid X, Z = 1] = \mathbb{E}[Y(0) \mid X, Z = 0]$ \label{a: cond_ign}
    \item (Population overlap)  $dQ/dP(x) < \infty$ for all $x \in \mathcal{X}$\label{a: pos}
\end{enumerate}
In our setup, we consider situations when the population overlap assumption \ref{a: pos} might be violated. In that case, researchers can instead
rely on parametric assumptions on $\mu(\x) = \mathbb{E}[Y(0) \mid X_i = \x]$, such as linearity, to identify and estimate the expected outcomes. 

Finally, following \citet{chattopadhyay2023implied}, we will focus on estimating the mean at a target \emph{covariate profile}, $\x^\star \in \mathcal{X}$, corresponding to our estimand of interest. For the ATT, this profile is simply the mean of the treated population, $\x^\star = \mathbb{E}[X \mid Z = 1]$, where $\mu(\x^\star) = \mathbb{E}[Y(0) \mid Z = 1]$.

\paragraph{Linear in features.}
Since we are focused on linear smoothers, we consider models that are linear in \textit{some} features, but which could be complex functions of the underlying covariates. This is an extremely large model class that ranges from simple linear models to the last layer embedding from a pre-trained large language model. For our setup, we let $\x$ be the features in the representation implied by the parametric model, rather than simply the raw covariates.
We further assume:
\begin{assumption}
    $\mu$ is H\"older continuous such that $|\mu(\x) - \mu(\x')| \leq a \cdot \|\x - \x'\|^\alpha$, with $a > 0$ and $\alpha > 0$.
\end{assumption}
Parameterizing $\mu$ in terms of its H\"older constants is useful for characterizing departures from linearity that directly affect the estimation error bound.

\paragraph{General linear estimands.} We can leverage recent work on the Riesz representer \citep{chernozhukov2022automatic} to immediately generalize our results to any linear functional of the data. Following \citet{bruns2023augmented}, for each unit $i \in [n]$, we observe the tuple $(X_i, Y_i, Z_i)$, with covariates $X_i \in \mathcal{X}$, outcome $Y_i \in \mathbb{R}$, and treatment $Z_i \in \mathcal{Z}$. The target functional is then 
$\mathbb{E}[h(X_i, Z_i, m)]$
for a function $h \in L_2$, where $m(x,z) = \mathbb{E}[Y_i \mid X_i = x, Z_i = z]$. 
Many common problems in causal inference and domain adaptation are special cases of this setup, including counterfactual quantities like the average derivative and the expected policy-specific outcome.
Finally, define a feature map $\phi : \mathcal{X} \times \mathcal{Z} \to \mathbb{R}^d$; the target \textit{feature profile} is then $\phi^*(x,z) = \mathbb{E}[ h(X, Z, \phi) ]$. Our results below apply by replacing the simple covariate profile $\x^\star$ with the much more general feature profile $\phi^\star(x,z)$.

\subsection{Weighting form of causal inference estimators}
\label{sec:weighting_intro}
Our focus is on \textit{weighting estimators} or \textit{linear smoothers} \citep{buja1989linear} of the form:
\begin{align*}
    \hat{\mu}(\x^\star) = \sum_{i = 1}^n w_{\star \leftarrow i}(\x_i) Y_i,
\end{align*}
with weights $w_{\star \leftarrow i}(\x_i)$, where $\star \leftarrow i$ emphasizes that the weights can depend both on the source covariates $\x_i$ and the target covariates $\x^\star$ \citep{lin2022regression}. When there is no ambiguity, we suppress the dependence on the covariates $\x_i$ and the target $\x^\star$.

A broad class of estimators have this form.
See \citet{knaus2024treatment} for a comprehensive discussion of the weighting form for common causal inference estimators. We highlight several special cases here, with a focus on whether the implied weights are constrained to be non-negative.

\paragraph{Explicit weighting estimators.} The first class of methods estimate the density ratio $\widehat{dQ/dP}(x)$, either directly or indirectly.
\begin{itemize}[leftmargin=*]
    \item
    \emph{Traditional Inverse Propensity Score Weighting.} In standard IPW \citep{rosenbaum1987model}, researchers first estimate a propensity score, $e(\x) = \mathbb{P}[Z_i = 1 \mid \X_i = \x]$ via a binary classifier like logistic regression, and then plug into a known functional form for $dQ/dP(x)$. For the ATT, $\hat{w}(\x) = \hat{e}(\x)/(1-\hat{e}(\x))$; since $\hat{e}(\x) \in (0,1)$, $\hat{w}_i(\x) > 0$ for all $i$.

    \item
    \emph{Balancing weights, synthetic control, and matching.} An alternative weighting approach instead directly estimates $dQ/dP(x)$ via constrained optimization \citep{ben2021balancing}. For example, consider the minimum variance weights that control imbalance in $\x$ between $P$ and $Q$:

\begin{equation}\label{eq:balancing_weight_optim}
    \hat{\mathbf{w}} \in \arg \min_{\mathbf{w} \in \mathcal{W}} \left\| \sum_{i=1}^n w_i \X_i - \x^\star\right\|^2_p + \lambda\|\w\|^2_2,
\end{equation}
where $\|\cdot\|_p$ is the $p$ vector norm and where $\mathcal{W}$ are possible constraints on the weights. \emph{Stable balancing weights} \citep{zubizarreta2015stable} and the \emph{Synthetic Control Method} \citep{abadie2010synthetic} are special cases where $\mathcal{W}$ is the simplex ($w_i \geq 0$, $\sum w_i = 1$) and the imbalance norm is $p = \infty$ and $p = 2$, respectively. \emph{Matching} is a special case where the weights are also constrained to be discrete.

    \item
    \emph{Riesz regression.} A final weighting approach, also known as automatic estimation of the Riesz representer \citep{chernozhukov2022debiased} also finds weights via Problem \eqref{eq:balancing_weight_optim}, albeit \textit{without} imposing the constraint that weights are non-negative. For example, minimum distance lasso Riesz regression in \citet{chernozhukov2022debiased} solves Equation \eqref{eq:balancing_weight_optim} with $\mathcal{W} = \mathbb{R}^n$ and $p = \infty$.
\end{itemize}

\paragraph{Linear smoothers and implicit weighting estimators.} A wide range of popular outcome models are linear smoothers that implicitly estimate weights $w$, including (kernel ridge) regression, $k$-nearest neighbors, random forests, xgboost, and many implementations of neural networks; see \citet{lin2022regression, curth2024random}. We highlight two prominent examples with and without a non-negativity constraint.

\begin{itemize}[leftmargin=*]
    \item \emph{(Kernel) ridge regression.} For features $\X$, the implied ridge regression weights are:
$$
w_{\star \leftarrow i} = {\x^\star}^\top (\X^\top\X + \lambda \mathbb{I})^{-1} \X_i,
$$
where $\lambda$ is a regularization parameter; ordinary least squares (OLS) as a special case when $\lambda = 0$. Kernel ridge regression is instead based on the implied kernel features $\phi(\x)$; see \citet{bruns2023augmented}, \citet{hirshberg2019minimax}. As \citet{bruns2023augmented} discuss, the ridge regression weights are equivalent to solving optimization problem \eqref{eq:balancing_weight_optim} with the imbalance norm set to $p = 2$ and with $\mathcal{W} = \mathbb{R}^n$, which does \emph{not} include a non-negativity constraint.

\item \emph{Random forests.} As \citet{athey2019generalized} discuss in the context of causal inference, (honest) random forests is a locally adaptive linear smoother with \textit{non-negative} weights:
$$
\hat{w}_{\star \leftarrow i} = \frac{1}{B} \sum_{b=1}^{B} \frac{\mathbb{I}\{\x^\star \in L_b(\x)\}}{|L_b(\x)|},
$$
where $L_b$ is the set of units that share a leaf node with the target $\x^\star$ and $b = 1, \ldots, B$ index the trees.
\end{itemize}

\paragraph{Augmented and hybrid estimators.} Finally, augmented or hybrid estimators combine initial weights $\w^0$ and outcome model $\hat{m}$:
\begin{align*}
    \hat\mu^{dr}(\x^\star) &=  \sum_{i=1}^N \hat{w}^0_i Y_i + \left( \hat{m}(\x^\star) - \sum_{i=1}^N w^0_i \hat{m}(\x_i) \right) \\&= \hat{m}(\x^\star) + \sum_{i=1}^N \hat{w}^0_i (Y_i - \hat{m}(\x_i)).
\end{align*}
When $\hat{m}$ is a linear smoother, then $\hat\mu^{dr}(\x)$ also has a weighting representation.
Let $\hat{m}(\x^\star) = \sum \hat{\omega}_i(\x) Y_i$ for a weighting function $\hat{\omega}: \mathbb{R}^d \to \mathbb{R}^n$. Following \citet{ben2021augmented}:
\begin{align*}
\hat{\mu}^{dr}(\x^\star)
&= \sum_{i=1}^N \left(\hat{w}^0_i + \hat{w}^{\text{adj}}_i\right)Y_i \\& \mbox{ where } \;\;\hat{w}^{\text{adj}}_i \equiv \hat{\omega}_i(\x^\star) - \sum_{j=1}^n \hat{w}^0_j \hat{\omega}_i(\x_j).
\end{align*}
For example, when the outcome model is ridge regression, the implied weights for the doubly robust estimator have the following form:
$$
\hat{w}^{dr}_i = \hat{w}^0_i + (\x^\star - \x'\hat{\w}^0)' (\x'\x + \lambda \mathbb{I})^{-1} \x_i.
$$
Importantly, even if the initial weights $\w^0$ are constrained to be non-negative, such as in traditional IPW, the implied doubly robust weights $\w^{dr}$ could be negative. In fact, the combined weights can be negative even if both the initial weights $\w^0$ and the outcome model-implied weights $\hat{\omega}$ are non-negative.

There are many examples of combined estimators of this form: standard Augmented IPW \citep{chattopadhyay2023implied}, bias correction for inexact matching \citep{lin2021nn_matching}, augmented synthetic control method \citep{ben2021augmented},  and regression-adjusted imputation estimators more broadly \citep{lin2022regression}. Finally, both debiased machine learning \citep{chernozhukov2018double} and \emph{automatic} debiased machine learning \citep{chernozhukov2022automatic} have this form. The former constrains the initial weights to be non-negative; the latter does not.

\section{REGULARIZING WORST-CASE EXTRAPOLATION BIAS}
Our goal is to bound the estimation error $\left| \mu(\x^\star) - \sum_{i=1}^n w_i Y_i \right|.$
We begin by building intuition for our approach in three steps.

\paragraph{Reflection representation.}
Under linearity, $\mu(-\x_i) = -\mu(\x_i)$, so a negative weight $w_i < 0$ on $\x_i$ yields $w_i \mu(\x_i) = |w_i| \mu(-\x_i)$; in other words, a negative weight $w_i$ is equivalent to applying a positive weight $|w_i|$ to the reflected point $-\x_i$. We can then construct a ``reflected'' estimator, denoted by $\ddagger$, which reflects points with negative weights around the origin:
\begin{align*}
\hat{\mu}^\ddagger (\x^\star) &= \sum_{i=1}^n w_i\ind(w_i \geq 0)
    \mu(\X_i) + |w_i|~\ind(w_i < 0) \mu(-\X_i) \\
    &= \sum_{i=1}^n |w_i| \mu({\X}^\ddagger_i), \qquad {\X}^\ddagger_i = \begin{cases}
        \X_i, & w_i \geq 0 \\
        -\X_i, & w_i < 0
    \end{cases},
\end{align*}
where $\hat\mu(\x^\star) = \hat\mu^\ddagger(\x^\star)$ if $\mu$ is an odd function, and where $w_i \X_i = |w_i| \X^\ddagger_i$ for all $i$.

\paragraph{Measuring parametric model violations.}
The difference between $\hat\mu(\x^\star)$ and $\hat\mu^\ddagger(\x^\star)$ measures the degree to which the assumed parametric model is violated. Specifically, we can write $\mu(-\x_i) = \delta(\x_i) - \mu(\x_i)$, where $\delta(\x_i) \equiv \mu(-\x_i) + \mu(\x_i)$. If $\mu$ is an odd function (e.g., $\mu$ is linear through the origin), then $\delta(\x_i) = 0$ for all $i$. Thus, $\delta(\x_i)$ is a point-specific measure of the extent to which the true outcome function violates the assumed parametric model; throughout, our discussion of ``nonlinearity'' should be understood as referring to such violations.

We use this representation to decompose the estimator $\hat{\mu}(\x^\star)$:
\begin{align*}
   &\hat{\mu}(\x^\star) = \sum_{i=1}^n w_i Y_i
    \;\;=\;\; \sum_{i = 1}^n \;w_i( \mu(\X_i) + \epsilon_i) \\
        &= \sum_{i=1}^n w_i\ind(w_i \geq 0)
    \mu(\X_i) + \\ 
    &\qquad \quad |w_i|~\ind(w_i < 0)
    \left(\mu(-\X_i) - \delta(\X_i)\right) + w_i\epsilon_i \\
    &= \underbrace{\sum_{i=1}^n |w_i| \mu({\X}^\ddagger_i)}_{\hat\mu^\ddagger(\x^\star)} + 
       \underbrace{\sum_{i=1}^n w_i \ind(w_i < 0) \delta(\X_i)}_{\text{model violation}} +  \underbrace{\sum_{i=1}^n w_i \epsilon_i}_{\text{noise}}.
\end{align*}

\paragraph{Error bound.}
Although $\delta(\X)$ is unknown, we can bound it via H\"older continuity: $|\delta(\x)| = |\mu(\x) + \mu(-\x)| \leq |\mu(\x) - \mu(\mathbf{0})| + |\mu(-\x) - \mu(\mathbf{0})| \leq 2a\|\x\|^\alpha$, where the last step uses $\mu(\mathbf{0}) = 0$, which holds after centering.\footnote{Replace $Y_i$ with $Y_i - \hat{\mu}(\mathbf{0})$, where $\hat{\mu}(\mathbf{0})$ is the fitted intercept.}
The resulting error bound is therefore
\begin{multline} \label{eq:error_bound}
\left|\mu(\x^\star) - \hat{\mu}(\x^\star)\right| \leq
\underbrace{\left|\sum_{i=1}^n |w_i|\mu({\X}^\ddagger_i) - \mu(\x^\star)\right|}_{\text{error in } \hat\mu^\ddagger(\x^\star)} \\
+ \underbrace{2a \sum_{i=1}^n |w_i|\ind(w_i < 0)  \|\X_i\|^\alpha}_{\text{error due to model violation}}
 + \underbrace{\left| \sum_{i=1}^n w_i\epsilon_i \right|}_{\text{noise}} .
\end{multline}
The first term directly depends on the imbalance between the target point $\x^\star$ and the re-weighted (reflected) training points $|\w|'\X^\ddagger$. The second term captures additional error due to model violation, corresponding to the $\delta(\X)$ term above; this is the key new term that our framework regularizes. The third term is the noise.

\subsection{Characterizing asymmetry-induced bias}
Thus far we have presented a conservative nonparametric bound. We now provide a slightly refined characterization by noting that the extent of the bias induced by negative weights is driven by the asymmetry in $\mu$. We do so by considering the decomposition of $\mu$ into its even and odd components, i.e., $\mu(x) = \mu_e(x) + \mu_o(x)$. By the definition of odd functions, we have $-\mu_o(x) = \mu_o(-x)$; we can then bound the worst-case risk of $\hat\w$ using the assumed H{\"o}lder constants $a$ and $\alpha$ and isolate the effect of the even component.
The formal statement is given below in Proposition~\ref{prop: err}; the proof is given in Appendix~\ref{sec: proof_err_bnd}.

\begin{proposition}
\label{prop: err}
    Let $\hat{\mu}(x^*) = \sum_{i=1}^n \hat{w}_i Y_i$ be the estimate of $\mu(x^*)$ with weights estimated via Equation~\eqref{eq: optimize} (defined below). Given $Y_i = \mu(X_i) + \epsilon_i$ where $\epsilon_i$ are independent random variables with $\mathbb{E}[\epsilon_i] = 0$ and finite second moment $\sigma^2 = \mathbb{E}[\epsilon_i^2]$, and $\mu$ is H\"older continuous with constants $a$ and $\alpha$. If $\epsilon_i$ are sub-Gaussian\footnote{We assume mean zero sub-Gaussian noise, analogous results can be obtained with this assumption replaced by bounded noise.} with parameter $\sigma$, then with probability at least $1-\delta$,
\begin{multline} \label{eq:prop_err_bound}
|\mu(x^*) - \hat{\mu}(x^*)| \leq \left|\mu(x^*) - \sum_{i=1}^n |\hat{w}_i|\,\mu(\X_i^\ddagger)\right| \\
+ 2\sum_{i=1}^n \lvert\hat{w}_i\rvert\,\mathbf{1}(\hat{w}_i < 0)\,a\|\X_i\|^\alpha + \sigma \|\hat{w}\|_2 \sqrt{2\log(2/\delta)}
\end{multline}
where $\X_i^\ddagger = \mathrm{sign}(\hat{w}_i)\X_i$ as defined above. The first term is the imbalance of the reflected estimator $\hat\mu^\ddagger(x^*) = \sum_i |\hat{w}_i|\mu(\X_i^\ddagger)$; the second is the model violation bias due to negative weights (identical to the second term of \eqref{eq:error_bound}); the third is the noise.
\end{proposition}
The proof in Appendix~\ref{sec: proof_err_bnd} proceeds via an even-odd decomposition that shows that negative weights introduce additional bias only through the \emph{even} component of $\mu$; the odd part is absorbed exactly into the first term via the reflection $\X_i^\ddagger$. The noise term in Proposition~\ref{prop: err} requires a sub-Gaussian assumption.

Since $\mu_e$ is unidentifiable from a single dataset, we construct a conservative worst-case form that does not require access to $\mu_e$. For completeness, Proposition~\ref{prop: erremp} in Appendix~\ref{sec: proof_err_bnd} provides an empirical analog that approximates $\mu_e$ via nearest-neighbor matching when the data are approximately symmetric.

Finally, following \citet{chattopadhyay2023implied}, we define \textit{negative influence} as the fraction of total weight on units with negative weights, $\sum_i\mathbf{1}[w_i < 0] |w_i| / \sum_i |w_i|$. This is a useful summary of the extent to which the estimate relies on extrapolation.

\subsection{Proposed Estimator}

We now propose an estimator to learn weights $\w$ that directly control the error bound in Equation \eqref{eq:error_bound}. To do so, we modify the standard balancing weights optimization problem in Equation \eqref{eq:balancing_weight_optim} by using the Lagrangian form of the non-negativity constraint, rather than the hard constraint. Thus, the combined estimator minimizes the error bound by controlling three terms: covariate imbalance, dispersion of the weights, and level of extrapolation:
\begin{align}\label{eq: optimize}
\hat{\mathbf{w}} \in \arg \min_{\mathbf{w}} \;\;
& \underbrace{\left\| \textstyle\sum_{i=1}^n w_i \X_i - \x^\star\right\|^2_2}_{(a)\;\text{imbalance}}
\;+\; \underbrace{\lambda \|\mathbf{w}\|^2_2}_{(b)\;\text{variance}} \notag \\
&+ \;\underbrace{\gamma \textstyle\sum_{i=1}^n \ind\!(w_i < 0)\, \lvert w_i\rvert \, \|\X_i\|^\alpha}_{(c)\;\text{extrapolation}}
\end{align}
where
\begin{itemize}[leftmargin=*]
    \item Term (a): Enforces balance between the target point \( \x^\star \) and the re-weighted training points \( \{\X_1, \dots, \X_n\} \), recalling that $w_i \X_i = |w_i| \X^\ddagger_i$ for all $i$. We focus on $p = 2$, but this generalizes to $p = \infty$. This corresponds to the first term of \eqref{eq:error_bound}.
    \item Term (b): Regularizes the dispersion of the weights \( \mathbf{w} \), controlling the noise term via $\|\hat{\mathbf{w}}\|_2$ in Proposition~\ref{prop: err}.
    \item Term (c): Penalizes model violation bias: the penalty $\gamma \sum_i \mathbf{1}(w_i<0)|w_i|\|\X_i\|^\alpha$ is proportional to the second term of \eqref{eq:error_bound}, with $\gamma$ scaling the sensitivity to parametric model violations.\footnote{In practice, $\alpha = 1$ corresponds to Lipschitz continuity; larger $\alpha$ assumes smoother departures from the parametric model and penalizes extrapolation less aggressively.}
\end{itemize}

Compared to the standard balancing weights problem \eqref{eq:balancing_weight_optim}, which trades off only imbalance and variance, the new objective \eqref{eq: optimize} introduces term (c) to control extrapolation. When the target lies outside the convex hull of the training points, achieving balance requires some weights to be negative, which increases both $\|\w\|_2$ and reliance on parametric assumptions. For $\gamma = 0$, Equation~\eqref{eq: optimize} recovers unconstrained balancing weights; at the other extreme, $\gamma \to \infty$ is equivalent to a hard non-negativity constraint. Increasing $\gamma$ reduces extrapolation bias and $\|\w\|_2$ but worsens imbalance in term (a). 

\textbf{Regularizing existing estimators.} Since many causal estimators have a weighting representation (Section~\ref{sec:weighting_intro}), we can regularize extrapolation in any baseline estimator with implied weights $\w'$ by solving
$$
\hat{\mathbf{w}} \in \arg \min_{\mathbf{w}} \|\mathbf{w} - \mathbf{w}' \|_2^2 + \gamma \textstyle\sum_{i=1}^n \ind(w_i < 0)\, \lvert w_i\rvert \, \|\X_i\|^\alpha.
$$
For example, the augmented synthetic control method \citep{ben2021augmented} first solves with non-negative weights, then augments with a ridge outcome model that implicitly introduces negative adjustment weights; the formulation above instead directly controls the degree of negativity through $\gamma$.

\textbf{Convexity.} Despite the indicator function in term (c), the optimization problem \eqref{eq: optimize} is strongly convex and admits a unique global minimizer. To see this, note that $\mathbf{1}(w_i < 0)|w_i| = \max(0, -w_i)$, which is convex as the pointwise maximum of two affine functions. Term (a) is a squared norm of an affine function of $\w$, hence convex, and term (b) is strongly convex with parameter $2\lambda$. More generally, term (c) can be written using an $\ell_p$ norm over the vector of per-unit penalties. When $p = 1$ (as written above and in Proposition~\ref{prop:regularized_bound}), introducing slack variables $s_i \geq -w_i$, $s_i \geq 0$ reduces the problem to a quadratic program (QP), which can be solved exactly in polynomial time with standard solvers. When $p = 2$, the problem becomes a second-order cone program (SOCP), which is likewise solvable in polynomial time.

Finally, we can specialize the error bound for our proposed estimator:

\begin{proposition}[Regularized Bound]
\label{prop:regularized_bound}
Let $\hat{w}$ solve Equation~\eqref{eq: optimize} and $\hat{\mu}(x^*) = \sum_{i=1}^n \hat{w}_i Y_i$ where $Y_i = \mu(X_i) + \epsilon_i$. Under the assumptions of Proposition~\ref{prop: err}, with probability at least $1 - \delta$:
\begin{multline*}
|\mu(x^*) - \hat{\mu}(x^*)| \leq \left|\mu(x^*) - \sum_{i=1}^n |\hat{w}_i|\,\mu(\X_i^\ddagger)\right| \\
+ 2\sum_{i=1}^n |\hat{w}_i|\,\mathbf{1}(\hat{w}_i < 0)\,a\|\X_i\|^\alpha + \sigma\|\x^*\|_2\sqrt{\log(2/\delta)/\lambda}
\end{multline*}
where $\X_i^\ddagger = \mathrm{sign}(\hat{w}_i)\X_i$ as in Proposition~\ref{prop: err}.
\end{proposition}

The proof is provided in the appendix. The variance term $\sigma\|\x^*\|_2\sqrt{\log(2/\delta)/\lambda}$ does not depend on $\gamma$: increasing $\gamma$ further constrains the feasible set and cannot inflate $\|\hat{w}\|_2$. This means regularizing extrapolation reduces the bias terms without incurring additional variance cost. The full bias-bias-variance tradeoff is demonstrated empirically in Sections~\ref{sec:synth} and~\ref{sec: moud}.

\subsection{Practical guidance: $\gamma$ as a sensitivity parameter}
We argue that $\gamma$ should be treated as a sensitivity parameter rather than a tuning parameter, and encourage researchers to examine the full set of estimates it spans. We recommend the following procedure:
\begin{enumerate}[leftmargin=*]
    \item Fix $\lambda$ via cross-validation on the standard balancing weights problem (i.e., with $\gamma = 0$).
    \item Sweep $\gamma$ over a grid from $0$ to a value $\gamma_{\max}$ at which all weights become non-negative.
    \item For each $\gamma$, record the point estimate, covariate balance (RMSE), and negative influence.
    \item Examine how estimates change across this range to assess sensitivity to parametric assumptions.
\end{enumerate}
If the goal is a single point estimate, we can instead choose $\gamma^*$ in the spirit of Lepski's method, selecting the largest $\gamma$ for which the change in the point estimate remains below a researcher-defined cutoff.

\section{SYNTHETIC DATA STUDY} \label{sec:synth}

\begin{figure}[t]
    \centering
    \includegraphics[width=0.30\textwidth]{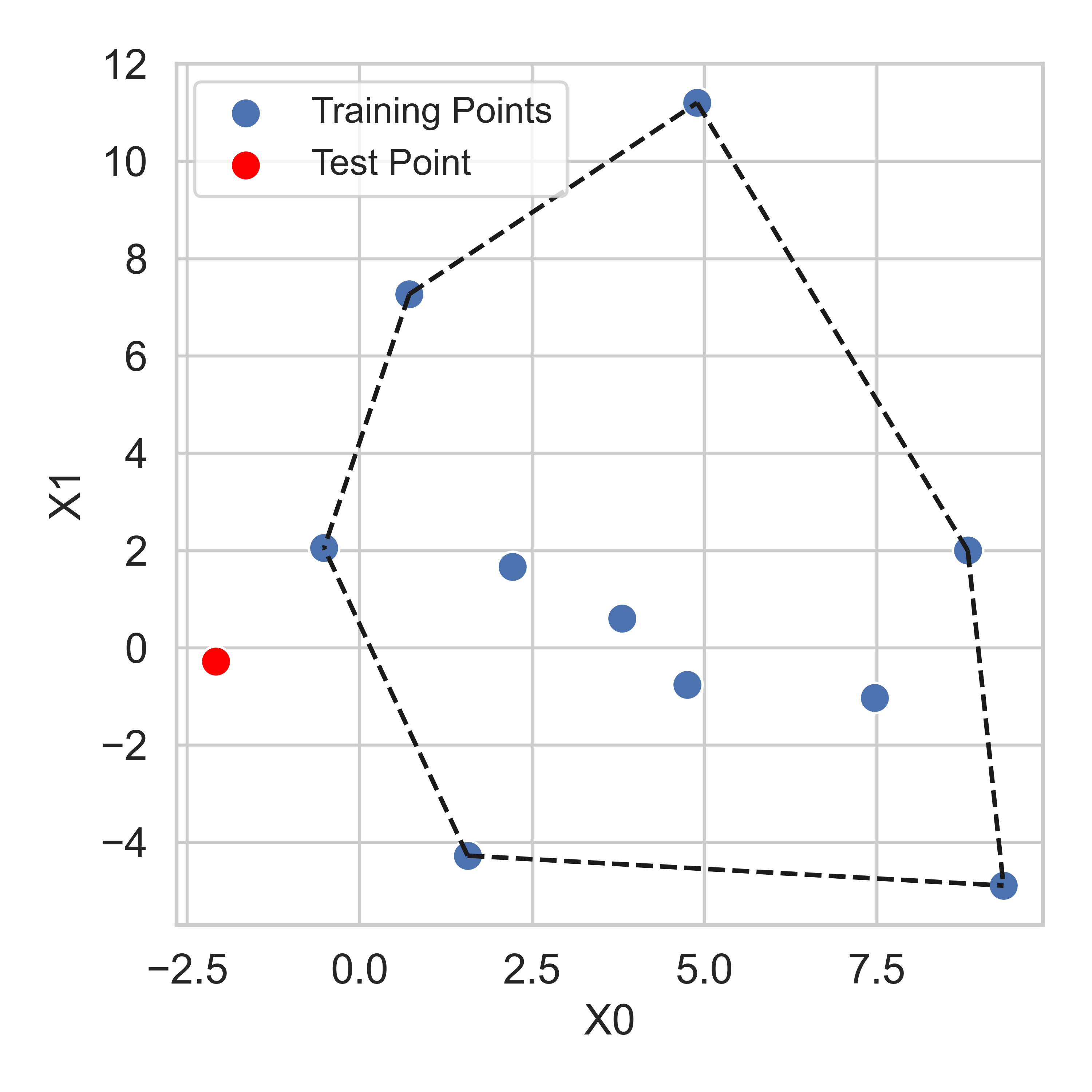}
    \caption{Convex hull of source (training) and target units. The target point lies outside the convex hull, requiring extrapolation.}
    \label{fig: convex_hull}
\end{figure}

We evaluate our approach using synthetic data with both linear and nonlinear data generating processes (DGPs) where the target point lies outside the convex hull of training points (Figure~\ref{fig: convex_hull}), creating a challenging extrapolation scenario with limited sample size ($n = 10$ training units, $n/p = 5$). We also consider a high-dimensional setting ($p = 5000$, $n/p = 0.2$) using the Friedman DGP. Full descriptions and additional figures are provided in Appendix~\ref{sec: simulation}.
 
\begin{figure}[t]
    \centering
    \includegraphics[width=0.75\linewidth]{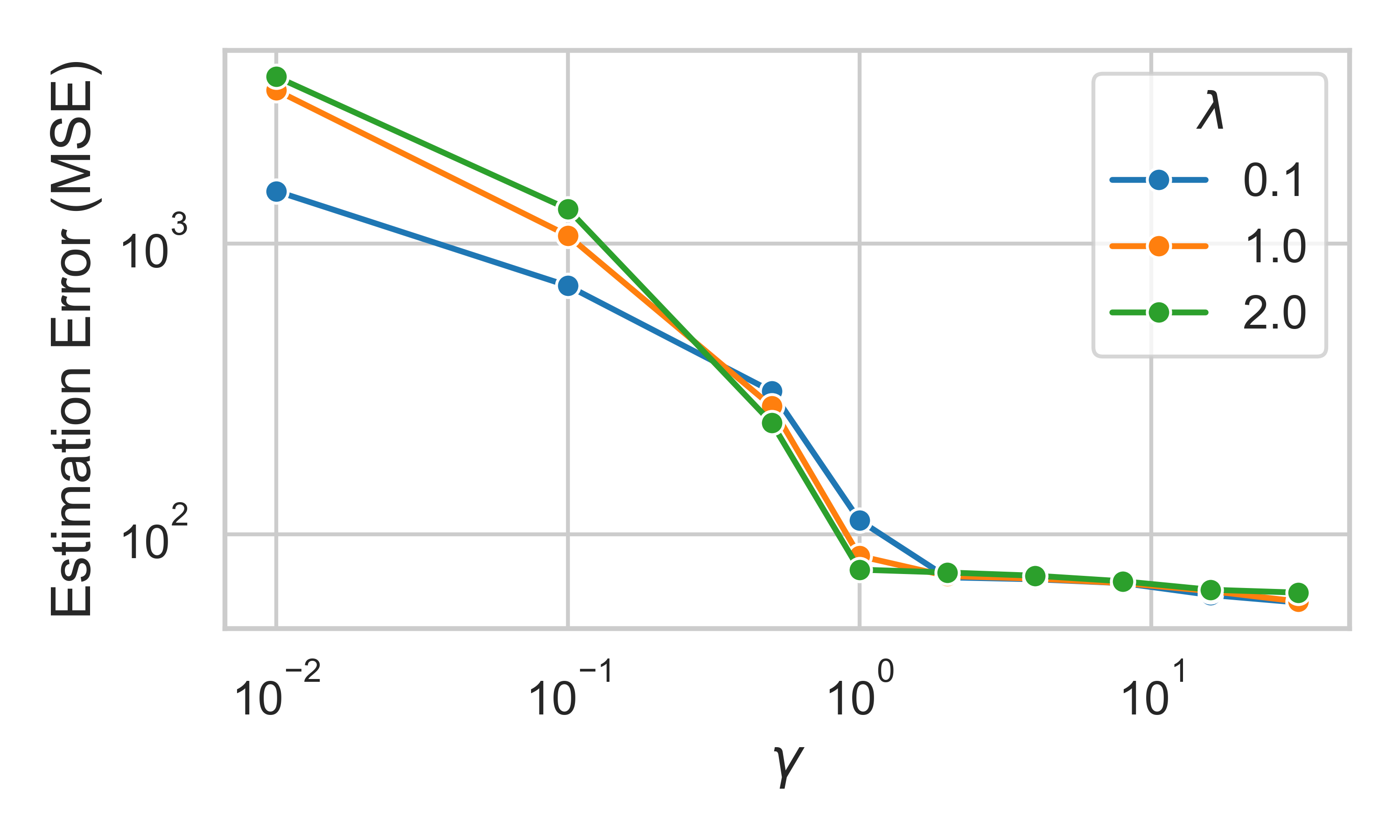}
    \caption{Estimation error (MSE) for the nonlinear DGP as a function of $\gamma$. The U-shaped curve illustrates the bias-bias-variance tradeoff: small $\gamma$ allows beneficial extrapolation, while large $\gamma$ incurs bias from poor balance.}
    \label{fig:sim_nonlinear_main}
\end{figure}

We consider two DGPs:
\begin{align*}
    \text{Linear:} \;\; \mu(\X) &= \beta^\top \X, \\
    \text{Nonlinear:} \;\; \mu(\X) &= 2X_1^2 + X_2 + X_1 X_2.
\end{align*}
For the linear DGP, where the parametric assumption holds, estimation error increases monotonically as we regularize extrapolation (i.e., increase $\gamma$): relying on correct parametric assumptions yields optimal estimates. However, for the nonlinear DGP (Figure~\ref{fig:sim_nonlinear_main}), the quadratic and interaction terms violate the linearity assumption. The results illustrate the bias-bias-variance tradeoff predicted by our theory: small amounts of extrapolation remain beneficial due to the linear component, but excessive extrapolation leads to high error rates due to violations of the parametric model. The high-dimensional experiment further demonstrates that sweeping over $\gamma$ smoothly interpolates between parametric regression-like behavior (unconstrained extrapolation) and IPW-like behavior (no extrapolation).

\section{GENERALIZING MEDICATION FOR OPIOID USE DISORDER TRIAL EVIDENCE} \label{sec: moud}
We now apply our framework to the problem of generalizing causal estimates from a randomized trial to a target population. Appendix~\ref{sec: transport_assumptions} states the formal identification assumptions and maps this setting onto the general framework of Section~\ref{sec:weighting_intro}.

The Starting Treatment With Agonist Replacement Therapies (START) trial ($S_i = 1$), initiated in 2006, was a multi-center study comparing buprenorphine versus methadone in treating opioid use disorder \citep{saxon2013buprenorphine, hser2014treatment}. The trial enrolled 1,271 participants, who were randomized in a 2:1 ratio to receive either buprenorphine or methadone. Methadone was found to have higher rates of patient retention in treatment compared to buprenorphine \citep{hser2014treatment}. Our analysis focuses on the outcome of relapse to regular opioid use within 24 weeks of medication assignment, defined as non-study opioid use for four consecutive weeks or daily use for seven consecutive days.

\citet{parikh2024we} identified that Hispanic women with a pre-treatment history of amphetamine and benzodiazepine use were underrepresented in the START trial relative to the target population ($S_i = 0$), highlighting a practical violation of the positivity assumption~\ref{a: positivity_transport}. In this study, we estimate the target average treatment effect (TATE), $\tau = \mathbb{E}[Y_i(1) - Y_i(0) \mid S_i = 0]$, for this underrepresented subgroup using our proposed framework alongside standard linear regression, gradient boosting regression (GBR), inverse probability weighting (IPW), and double machine learning estimators.

The target sample is drawn from the 2015–2017 Treatment Episode Dataset - Admissions (TEDS-A), which includes data on individuals entering publicly funded substance use treatment programs across 48 states (excluding Oregon and Georgia) and the District of Columbia. Our analysis focuses on Hispanic women with a pre-treatment history of amphetamine and benzodiazepine use.

We code methadone as $Z=1$ and buprenorphine as $Z=0$, with $Y=1$ representing relapse. Pretreatment covariates include age, race, biological sex, and substance use history (amphetamine, benzodiazepines, cannabis, and intravenous drug use) measured at the initiation of medication for opioid use disorder (MOUD) treatment. For each treatment arm $z \in \{0,1\}$, we estimate $\mu_z(\x^\star) = \mathbb{E}[Y_i \mid X_i = \x^\star, S_i = 1, Z_i = z]$ at the target profile $\x^\star = \mathbb{E}[X_i \mid S_i = 0]$ using the trial participants assigned to arm $z$ as source units. The estimated TATE is then $\hat{\tau} = \hat{\mu}_1(\x^\star) - \hat{\mu}_0(\x^\star)$.

\begin{figure}
    \centering
    \includegraphics[width=\linewidth]{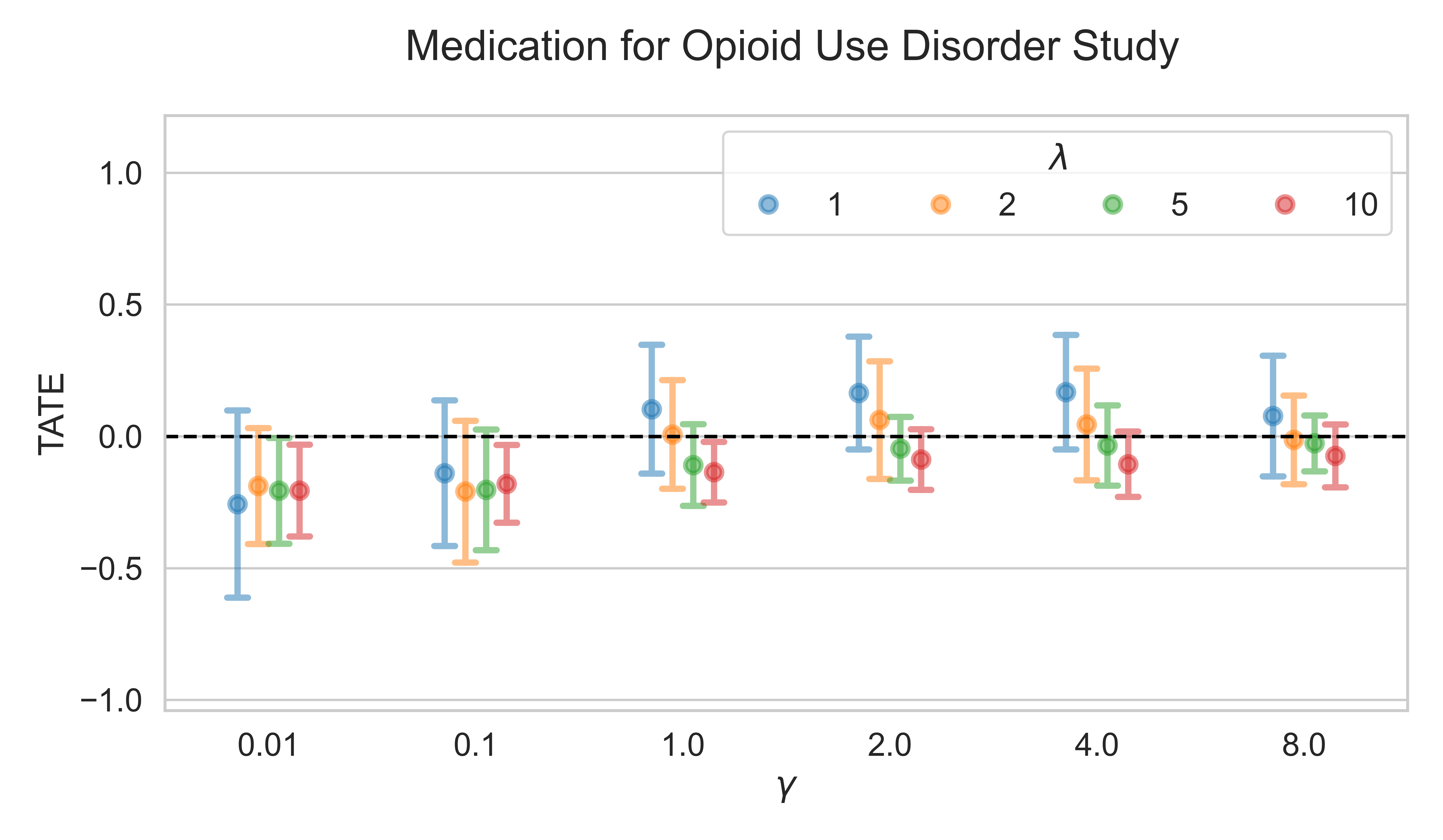}\\
    \includegraphics[width=0.6\linewidth]{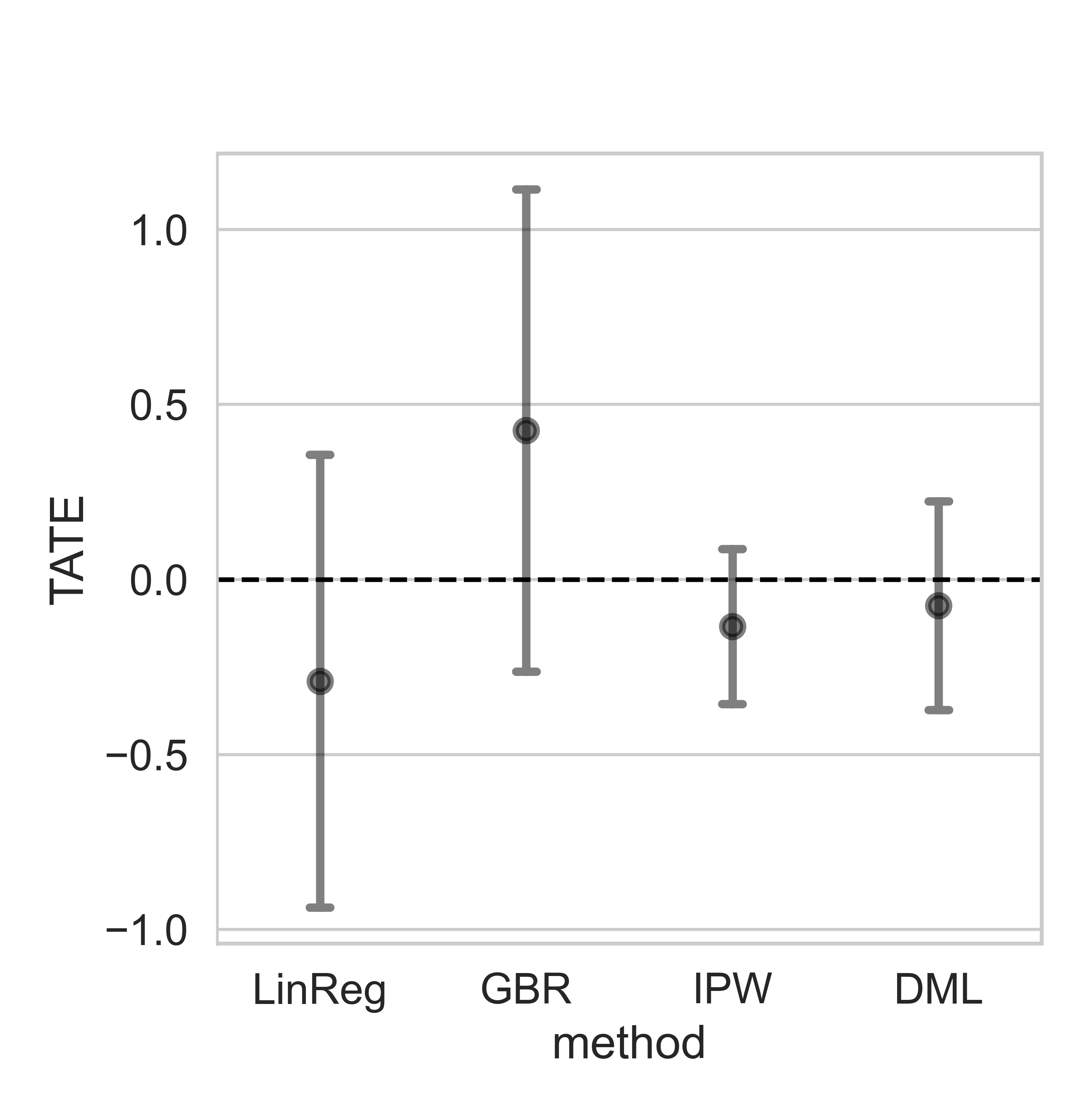}
    \caption{Target Average Treatment Effects for the Target Sample for Hispanic Females who have a history of Amphetamine and Benzodiazepine use in TEDS-A population. Each hue corresponds to a value of $\lambda$ and the x-axis corresponds to different values of $\gamma$ (on log scale).}
    \label{fig:moud_ate}
\end{figure}

\begin{figure}
    \centering
    \includegraphics[width=\linewidth]{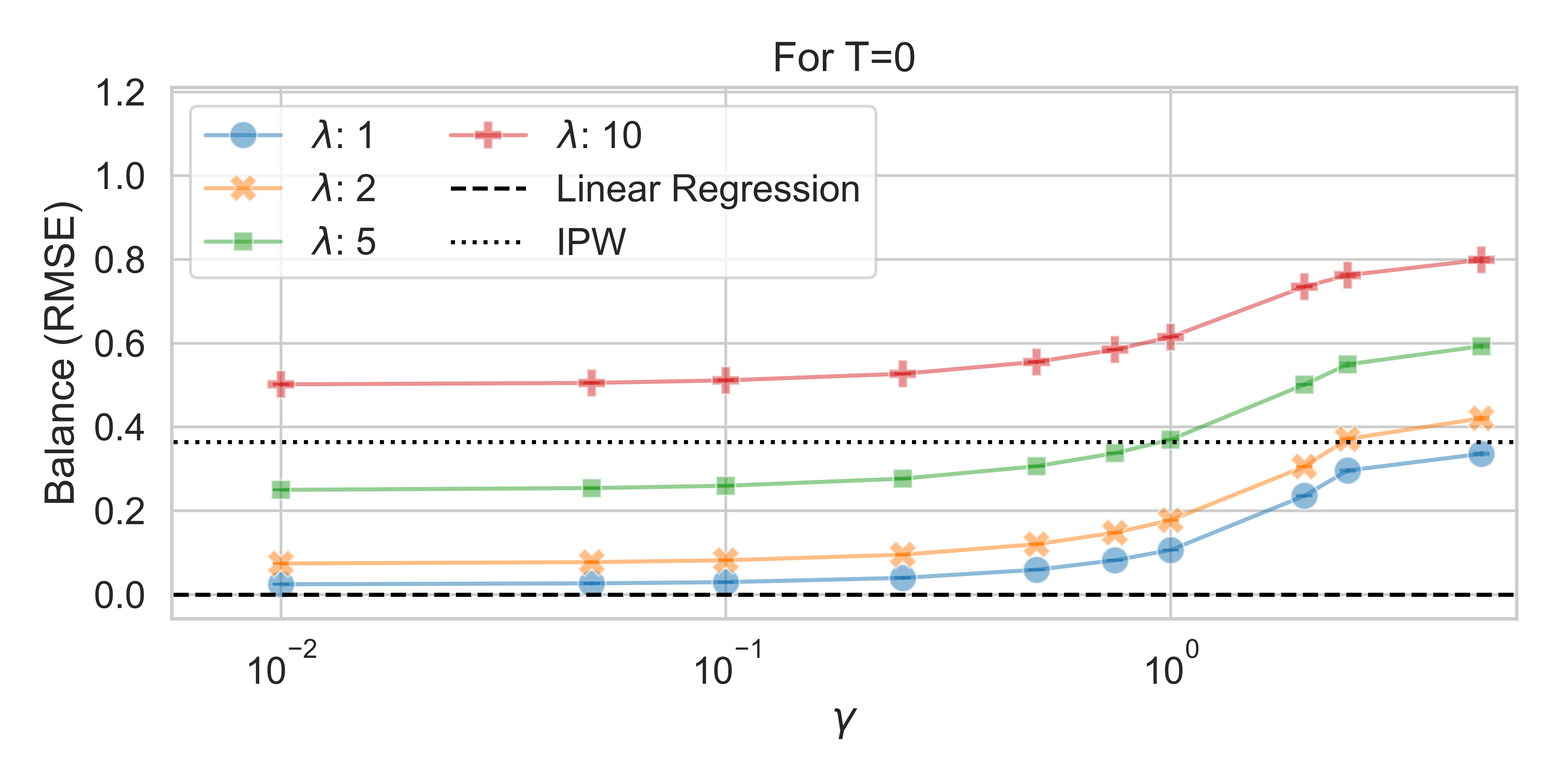}\\
    \includegraphics[width=\linewidth]{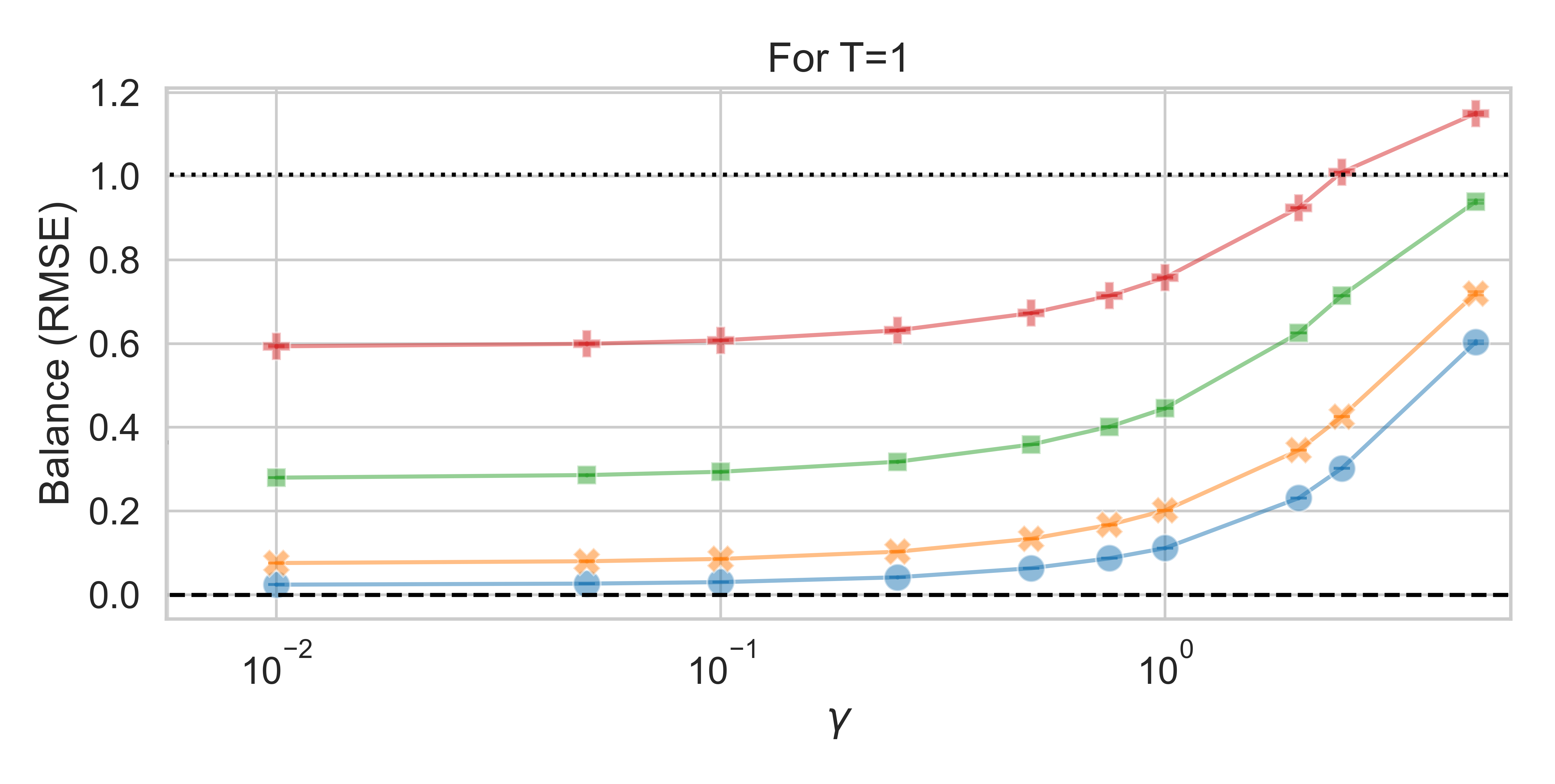}\\
    \caption{Balance between the trial and the target samples measured as the root mean squared error (RMSE) for different values of $\gamma$ and $\lambda$.}
    \label{fig:moud_diagnostics_balance}
\end{figure}

\begin{figure}
    \centering
    \includegraphics[width=\linewidth]{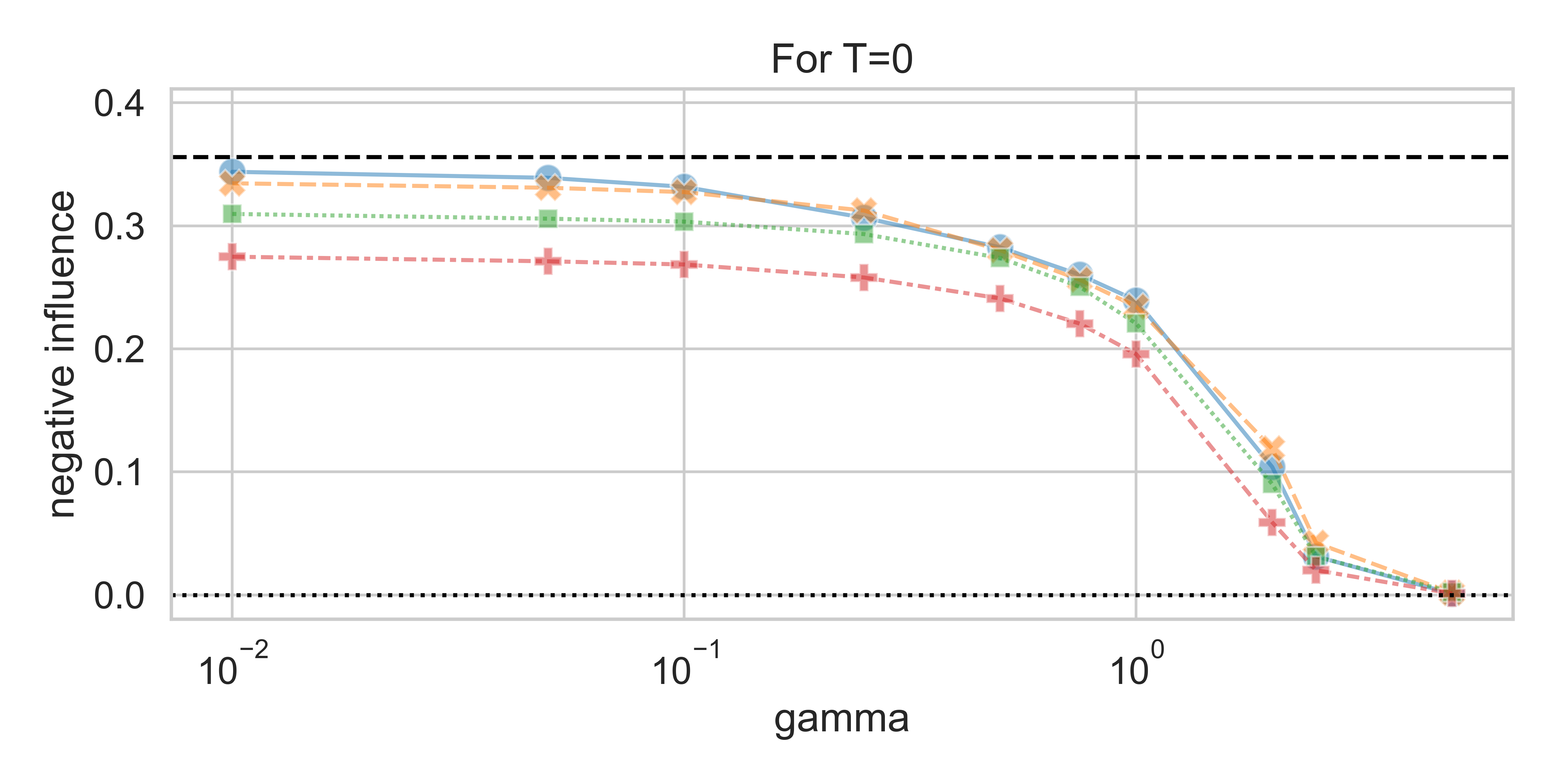}\\
    \includegraphics[width=\linewidth]{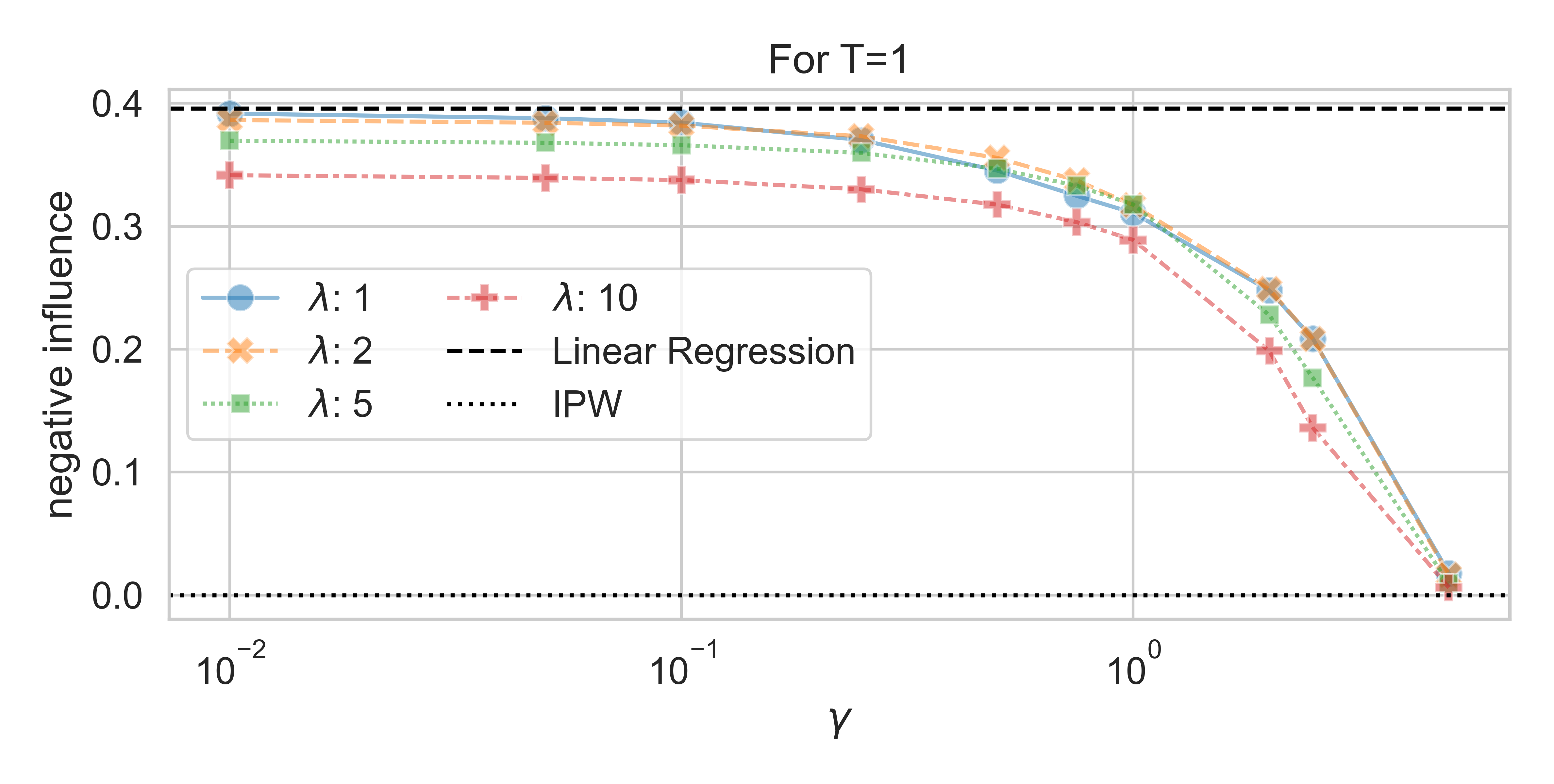}
    \caption{Negative influence, defined as the contribution of negative weights in estimation, for different values of $\gamma$ and $\lambda$.}
    \label{fig:moud_diagnostics_neg_inf}
\end{figure}

We then apply our proposed framework to this problem. By varying \(\gamma\) from 0.01 to 10, we examine how treatment effect estimates shift with increasing regularization of negative weights. Without regularization, the point estimates converge to those from linear regression. As regularization intensifies, however, the estimates smoothly shift towards zero and occasionally change sign from negative to positive for smaller values of \(\lambda\). This sensitivity underscores the influence of assumptions on the point estimates. While increasing \(\gamma\) reduces negative influence (Figure~\ref{fig:moud_diagnostics_neg_inf}), it worsens covariate balance, as reflected in higher RMSE values (Figure~\ref{fig:moud_diagnostics_balance}). Thus, our framework highlights a trade-off between minimizing reliance on parametric assumptions and achieving optimal covariate balance. Applied researchers should therefore interpret treatment effect estimates for this under-represented subgroup with caution given the sensitivity to modeling assumptions. As \citet{parikh2024we} emphasized, collecting more representative trial data is critical to credibly estimate treatment effects.

\section{CONCLUSION}
This work proposes a framework for regularizing extrapolation in causal inference by replacing hard non-negativity constraints with soft penalties on negative weights. Our theoretical error bounds show a fundamental ``bias-bias-variance'' tradeoff between distributional imbalance, model misspecification, and estimator variance, decomposing extrapolation bias through a novel reflection perspective. Empirically, synthetic data experiments confirm that controlled extrapolation smoothly interpolates between fully constrained and unconstrained approaches. A real-world medication trial illustrates how sweeping over the regularization parameter provides a practical sensitivity analysis for transportability estimates under positivity violations.

\textbf{Limitations and Future Work.} Our approach focuses on weighting-type estimators and relies on H\"{o}lder continuity and conditional ignorability, which may not hold in practice. Operationalizing sensitivity analysis for unmeasured confounding is a critical next step; existing proposals for balancing weights \citep{soriano2023interpretable} do not directly apply to our framework, and adapting such methods is an important direction. More broadly, future work should extend the bias-bias-variance tradeoff analysis to more flexible estimator classes and weaker continuity assumptions. A key open question is to characterize data-generating processes under which soft-constrained extrapolation ($\gamma > 0$) provably improves MSE relative to both unconstrained ($\gamma = 0$) and fully constrained ($\gamma \to \infty$) estimators, for example when the density ratio is large near $\x^\star$. Finally, our theoretical results assume sub-Gaussian noise, though analogous bounds follow under bounded outcomes via Hoeffding-type inequalities.

\section*{ACKNOWLEDGMENTS}
The authors would like to thank the reviewers, the area chair, and the program chair of AISTATS 2026 for their constructive input to help improve the paper. Harsh Parikh, Kara Rudolph, and Elizabeth Stuart would like to acknowledge that this work was funded by NIH NIDA R01DA056407.

\bibliographystyle{abbrvnat}
\bibliography{references}

\section*{CHECKLIST}

\begin{enumerate}

  \item For all models and algorithms presented, check if you include:
  \begin{enumerate}
    \item A clear description of the mathematical setting, assumptions, algorithm, and/or model. [\textbf{Yes}]
    \item An analysis of the properties and complexity (time, space, sample size) of any algorithm. [\textbf{Yes}]
    \item (Optional) Anonymized source code, with specification of all dependencies, including external libraries. [\textbf{No}]
  \end{enumerate}

  \item For any theoretical claim, check if you include:
  \begin{enumerate}
    \item Statements of the full set of assumptions of all theoretical results. [\textbf{Yes}]
    \item Complete proofs of all theoretical results. [\textbf{Yes}]
    \item Clear explanations of any assumptions. [\textbf{Yes}]     
  \end{enumerate}

  \item For all figures and tables that present empirical results, check if you include:
  \begin{enumerate}
    \item The code, data, and instructions needed to reproduce the main experimental results (either in the supplemental material or as a URL). [\textbf{Yes}]
    \item All the training details (e.g., data splits, hyperparameters, how they were chosen). [\textbf{Yes}]
    \item A clear definition of the specific measure or statistics and error bars (e.g., with respect to the random seed after running experiments multiple times). [\textbf{Yes}]
    \item A description of the computing infrastructure used. (e.g., type of GPUs, internal cluster, or cloud provider). [\textbf{Yes}]
  \end{enumerate}

  \item If you are using existing assets (e.g., code, data, models) or curating/releasing new assets, check if you include:
  \begin{enumerate}
    \item Citations of the creator If your work uses existing assets. [\textbf{Yes}]
    \item The license information of the assets, if applicable. [\textbf{Not Applicable}]
    \item New assets either in the supplemental material or as a URL, if applicable. [\textbf{Not Applicable}]
    \item Information about consent from data providers/curators. [\textbf{Not Applicable}]
    \item Discussion of sensible content if applicable, e.g., personally identifiable information or offensive content. [\textbf{Not Applicable}]
  \end{enumerate}

  \item If you used crowdsourcing or conducted research with human subjects, check if you include:
  \begin{enumerate}
    \item The full text of instructions given to participants and screenshots. [\textbf{Not Applicable}]
    \item Descriptions of potential participant risks, with links to Institutional Review Board (IRB) approvals if applicable. [\textbf{Not Applicable}]
    \item The estimated hourly wage paid to participants and the total amount spent on participant compensation. [\textbf{Not Applicable}]
  \end{enumerate}

\end{enumerate}

\clearpage
\appendix
\thispagestyle{empty}

\onecolumn
\aistatstitle{Regularizing Extrapolation in Causal Inference:\\Supplementary Materials}

\section{IMPLEMENTATION DETAILS}
We implement the methods and case studies in this paper using Python 3.10. We implemented our weight estimation framework using PyTorch (version 2.7.0) for efficient automatic differentiation and optimization. The optimization is performed using Adam optimizer with a default learning rate of 0.01 for a default of 5,000 epochs. By default the weights are normalized to sum to one after each optimization step -- however, user can choose otherwise. The implementation includes comprehensive visualization tools, including Love plots for covariate balance assessment and 2D scatter plots with convex hull visualization for geometric interpretation. All random operations are seeded for reproducibility, and the code supports both single and multiple outcome variables. For the MOUD case study in Section~\ref{sec: moud}, we scale the pre-treatment data to ensure that the maximum value for each covariate is 1 and the minimum is 0; it is important to note that most covariates in this instance are discrete binary.
\textit{Computational details:} Our objective mimics a classic quadratic program (QP). QPs are well studied in the literature, and their run time complexity is known to be $O(n^{3.67} * log n)$ where $n$ is the number of training units \citep{kapoor1986fast}. Our new simulation study also demonstrates that our method is scalable to high-dimensional data.

\section{TRANSPORTABILITY ASSUMPTIONS FOR THE MOUD APPLICATION}
\label{sec: transport_assumptions}

The main text develops the framework for estimating $\mu(\x^\star) = \mathbb{E}[Y(0) \mid X = \x^\star]$ in the ATT setting. The MOUD application in Section~\ref{sec: moud} instead concerns \emph{transportability}: generalizing treatment effect estimates from a randomized trial to a target population. Here we state the additional assumptions required and show how the general framework applies. For a comprehensive review, see \citet{degtiar2023review}.

\paragraph{Setup.} Let $S_i \in \{0, 1\}$ indicate membership in the trial ($S_i = 1$) versus the target population ($S_i = 0$), and let $Z_i \in \{0, 1\}$ denote treatment assignment within the trial. The estimand is the \emph{target average treatment effect} (TATE):
\[
\tau = \mathbb{E}[Y_i(1) - Y_i(0) \mid S_i = 0].
\]
Because neither potential outcome is observed in the target population, estimating the TATE requires identifying both $\mathbb{E}[Y_i(z) \mid S_i = 0]$ for $z \in \{0, 1\}$. We require the following assumptions:
\begin{enumerate}[label=T.\arabic*., leftmargin=*]
    \item \label{a: sutva_transport} (SUTVA) $Y_i = Y_i(z)$ when $Z_i = z$.
    \item \label{a: internal_validity} (Ignorability and overlap within the trial) $Z_i \perp\!\!\!\perp \{Y_i(0), Y_i(1)\} \mid X_i,\, S_i = 1$ and $\Pr(Z_i = z \mid X_i = \x,\, S_i = 1) > 0$. Both hold by design via randomization.
    \item \label{a: mean_exchange} (Mean exchangeability over populations) For each $z \in \{0,1\}$:
    \[
    \mathbb{E}[Y_i(z) \mid X_i = \x,\, S_i = 1] = \mathbb{E}[Y_i(z) \mid X_i = \x,\, S_i = 0].
    \]
    This assumes that, conditional on covariates, expected potential outcomes do not differ between the trial and target populations.
    \item \label{a: positivity_transport} (Positivity of trial enrollment) $\Pr(S_i = 1 \mid X_i = \x) > 0$ for all $\x$ in the support of the target population.
\end{enumerate}

As with the main results, a key goal is to assess sensitivity to violations of this last positivity assumption \ref{a: positivity_transport}.

\paragraph{Connection to the general framework.} Under \ref{a: sutva_transport}--\ref{a: mean_exchange}, for each treatment arm $z$:
\[
\mathbb{E}[Y_i(z) \mid S_i = 0] = \mathbb{E}\!\left[\mu_z(X_i) \mid S_i = 0\right],
\]
where $\mu_z(\x) \equiv \mathbb{E}[Y_i \mid X_i = \x,\, S_i = 1,\, Z_i = z]$ is the conditional mean outcome among trial participants assigned to treatment $z$. This is precisely the estimation problem in Section~\ref{sec:weighting_intro}: the source units are trial participants in arm $z$ (with observed outcomes), $\mu_z$ plays the role of $\mu$, and the target covariate profile is $\x^\star = \mathbb{E}[X_i \mid S_i = 0]$. Assumption~\ref{a: positivity_transport} is the analog of the population overlap assumption \ref{a: pos}; its violation for Hispanic women with amphetamine and benzodiazepine use history is the central challenge addressed in Section~\ref{sec: moud}. The TATE is then $\tau = \hat{\mu}_1(\x^\star) - \hat{\mu}_0(\x^\star)$, where each arm is estimated separately using our proposed weights.

\section{PROOFS}
\label{sec: proof_err_bnd}
\subsection{Proof of Proposition \ref{prop: err}}
\begin{proof}
First taking the bound of the estimation error
\begin{align*}
|\mu(x^*) - \hat{\mu}(x^*)| &= \left|\mu(x^*) - \sum_{i=1}^n \hat{w}_i Y_i\right| \\
&= \left|\mu(x^*) - \sum_{i=1}^n \hat{w}_i (\mu(X_i) + \epsilon_i)\right| \\
&\leq \left|\mu(x^*) - \sum_{i=1}^n \hat{w}_i \mu(X_i)\right| + \left|\sum_{i=1}^n \hat{w}_i \epsilon_i\right|
\end{align*}

We bound the bias term $\left|\mu(x^*) - \sum_{i=1}^n \hat{w}_i \mu(X_i)\right|$ by decomposing $\mu = \mu_e + \mu_o$ and establishing two preliminary identities.

\textbf{Odd component.} Since $\mu_o$ is odd, $\mathrm{sign}(w)\mu_o(X) = \mu_o(\mathrm{sign}(w)X)$ for any scalar $w$ and vector $X$. Writing $\hat{w}_i = |\hat{w}_i|\,\mathrm{sign}(\hat{w}_i)$ and recalling $X_i^\ddagger = \mathrm{sign}(\hat{w}_i)X_i$,
\begin{align*}
\sum_{i=1}^n \hat{w}_i \mu_o(X_i)
= \sum_{i=1}^n |\hat{w}_i|\,\mathrm{sign}(\hat{w}_i)\,\mu_o(X_i)
= \sum_{i=1}^n |\hat{w}_i|\,\mu_o\!\left(\mathrm{sign}(\hat{w}_i)X_i\right)
= \sum_{i=1}^n |\hat{w}_i|\,\mu_o(X_i^\ddagger),
\end{align*}
so $\sum_{i=1}^n \hat{w}_i \mu_o(X_i) = \sum_{i=1}^n |\hat{w}_i|\,\mu_o(X_i^\ddagger)$ exactly.

\textbf{Even component.} Since $\mu_e$ is even, $\mu_e(X_i^\ddagger) = \mu_e(\mathrm{sign}(\hat{w}_i)X_i) = \mu_e(X_i)$ for all $i$. Applying H\"older continuity and the centering condition $\mu_e(\mathbf{0}) = 0$ (which holds after centering as described in the main text):
\begin{align*}
|\mu_e(X_i)| = |\mu_e(X_i) - \mu_e(\mathbf{0})| \leq a\|X_i\|^\alpha.
\end{align*}

\textbf{Combining.} The odd-component analysis gives $\sum_i \hat{w}_i\mu_o(X_i) = \sum_i |\hat{w}_i|\mu_o(X_i^\ddagger)$, and writing $\hat{w}_i = |\hat{w}_i| - 2|\hat{w}_i|\mathbf{1}(\hat{w}_i<0)$ for the even component gives
\begin{align*}
\sum_i \hat{w}_i\mu(X_i) &= \sum_i |\hat{w}_i|\mu_e(X_i) - 2\sum_i |\hat{w}_i|\mathbf{1}(\hat{w}_i<0)\mu_e(X_i) + \sum_i |\hat{w}_i|\mu_o(X_i^\ddagger) \\
&= \sum_{i=1}^n |\hat{w}_i|\mu(X_i^\ddagger) - 2\sum_{i=1}^n |\hat{w}_i|\mathbf{1}(\hat{w}_i < 0)\mu_e(X_i),
\end{align*}
where the last step uses $\mu_e(X_i^\ddagger) = \mu_e(X_i)$. The triangle inequality then gives
\begin{align*}
|\mu(x^*) - \hat{\mu}(x^*)| \leq \left|\mu(x^*) - \sum_{i=1}^n |\hat{w}_i|\,\mu(X_i^\ddagger)\right| + 2\sum_{i=1}^n |\hat{w}_i|\,\mathbf{1}(\hat{w}_i < 0)\,|\mu_e(X_i)| + \left|\sum_{i=1}^n \hat{w}_i \epsilon_i\right|.
\end{align*}
Applying $|\mu_e(X_i)| \leq a\|X_i\|^\alpha$ from the even-component analysis above:
\begin{align*}
|\mu(x^*) - \hat{\mu}(x^*)| \leq \left|\mu(x^*) - \sum_{i=1}^n |\hat{w}_i|\,\mu(X_i^\ddagger)\right| + 2\sum_{i=1}^n |\hat{w}_i|\,\mathbf{1}(\hat{w}_i < 0)\,a\|X_i\|^\alpha + \left|\sum_{i=1}^n \hat{w}_i \epsilon_i\right|.
\end{align*}

Now turning to the noise term, we have by assumption that the sum $\sum_{i=1}^n \hat{w}_i \epsilon_i$ is sub-Gaussian with parameter $\sigma \|\hat{w}\|_2$. Using standard sub-Gaussian concentration,
\begin{align*}
P\left(\left|\sum_{i=1}^n \hat{w}_i \epsilon_i\right| > t \,\Big|\, \hat{w}\right) \leq 2\exp\left(-\frac{t^2}{2\sigma^2 \|\hat{w}\|_2^2}\right)
\end{align*}
Solving for $t$, setting the right hand side to $\delta$:
\begin{align*}
t^2 = 2\sigma^2 \|\hat{w}\|_2^2 \log\left(\frac{2}{\delta}\right),
\end{align*}
so $t = \sigma \|\hat{w}\|_2 \sqrt{2\log(2/\delta)}$, and with probability at least $1-\delta$,
\begin{align*}
\left|\sum_{i=1}^n \hat{w}_i \epsilon_i\right| \leq \sigma \|\hat{w}\|_2 \sqrt{2\log(2/\delta)}.
\end{align*}
The desired statement follows by combining the bias and noise bounds.
\end{proof}

\subsection{Proposition \ref{prop: erremp} and Proof}

\begin{proposition}[Approximate Bounds]
\label{prop: erremp}
Let $\hat{\mu}(x^*) = \sum_{i=1}^n \hat{w}_i Y_i$ be the estimate of $\mu(x^*)$ with weights estimated via Equation~\ref{eq: optimize}. Given $Y_i = \mu(X_i) + \epsilon_i$ where $\epsilon_i$ are independent sub-Gaussian random variables with parameter $\sigma$, $\mu$ is H\"older continuous with constants $a$ and $\alpha$, and $\sum_{i=1}^n \hat{w}_i = 1$. Let $I_{\text{paired}} = \{i : -X_i \in \{X_1, \ldots, X_n\}\}$, $I_{\text{nn}} = \{1,\ldots,n\} \setminus I_{\text{paired}}$, and for each $i \in I_{\text{nn}}$, define $j^*(i) = \arg\min_{j \neq i} \|X_j - (-X_i)\|$. Then with probability at least $1-\delta$, $|\mu(x^*) - \hat{\mu}(x^*)|$ is at most
\begin{align*}
&\Big|\textstyle\sum_{i \in I_{\text{paired}}} \hat{w}_i\frac{Y_i + Y_{-i}}{2} + \textstyle\sum_{i \in I_{\text{nn}}} \hat{w}_i\frac{Y_i + Y_{j^*(i)}}{2} - \mu_e(x^*) \textstyle\sum_{i=1}^n \hat{w}_i\Big| \\
&\quad + \textstyle\sum_{i \in I_{\text{nn}}} |\hat{w}_i|\, a\|X_{j^*(i)} - (-X_i)\|^\alpha \\
&\quad + 2\textstyle\sum_{i=1}^n |\hat{w}_i|\mathbf{1}(\hat{w}_i < 0)\,a\|X_i - x^*\|^\alpha \\
&\quad + \sigma\|\hat{w}\|_2\sqrt{2\log(6/\delta)} \\
&\quad + \tfrac{\sigma}{\sqrt{2}}\bigl(\|\hat{w}_{I_{\text{paired}}}\|_2 + \|\hat{w}_{I_{\text{nn}}}\|_2\bigr)\sqrt{2\log(6/\delta)}
\end{align*}
where $\mu_e(x^*)$ is bounded using the closest observation $j^* = \arg\min_{j} \|X_j - x^*\|$:
\begin{align*}
|\mu_e(x^*)| \leq\;
&\left|\frac{Y_{j^*} + Y_{j^*(j^*)}}{2}\right| + \sigma\sqrt{2\log(6/\delta)} \\
&\quad + a\|X_{j^*(j^*)} - (-X_{j^*})\|^\alpha + a\|X_{j^*} - x^*\|^\alpha.
\end{align*}
\end{proposition}

\begin{proof}
Our approach will be to construct an observable bound by replacing $\mu_e(X_i)$ and $\mu_e(x^*)$ with empirical estimates. Starting from the bias-variance decomposition and applying the even-odd decomposition (as in the proof of Proposition~\ref{prop: err}), the estimation error is bounded by
\begin{align*}
|\mu(x^*) - \hat{\mu}(x^*)| \leq \left|\sum_{i=1}^n \hat{w}_i[\mu_e(X_i) - \mu_e(x^*)]\right| + 2\sum_{i=1}^n |\hat{w}_i|\mathbf{1}(\hat{w}_i < 0)\,a\|X_i\|^\alpha + \left|\sum_{i=1}^n \hat{w}_i \epsilon_i\right|.
\end{align*}

We first rewrite the bias as $$\left|\sum_{i=1}^n \hat{w}_i \mu_e(X_i) - \mu_e(x^*) \sum_{i=1}^n \hat{w}_i\right|$$
and replace $\mu_e(X_i)$ with observable approximations.
For $i \in I_{\text{paired}}$:, $$\mu_e(X_i) = \frac{Y_i + Y_{-i}}{2} - \frac{\epsilon_i + \epsilon_{-i}}{2}$$. For $i \in I_{\text{nn}}$, $$\mu_e(X_i) = \frac{Y_i + Y_{j^*(i)}}{2} - \frac{\epsilon_i + \epsilon_{j^*(i)}}{2} + \frac{\mu(X_{j^*(i)}) - \mu(-X_i)}{2}$$ where $Y_{j^*(i)}$ is the approximation of $Y_{-i}$ using NN, and $\left|\frac{\mu(X_{j^*(i)}) - \mu(-X_i)}{2}\right| \leq a\|X_{j^*(i)} - (-X_i)\|^\alpha$ by H\"older continuity.
Substituting those terms in gives
\begin{align*}
&\left|\sum_{i=1}^n \hat{w}_i \mu_e(X_i) - \mu_e(x^*) \sum_{i=1}^n \hat{w}_i\right| \leq \left|\sum_{i \in I_{\text{paired}}} \hat{w}_i\frac{Y_i + Y_{-i}}{2} + \sum_{i \in I_{\text{nn}}} \hat{w}_i\frac{Y_i + Y_{j^*(i)}}{2} - \mu_e(x^*) \sum_{i=1}^n \hat{w}_i\right| \\&+ \left|\sum_{i \in I_{\text{paired}}} \hat{w}_i\frac{\epsilon_i + \epsilon_{-i}}{2}\right| + \left|\sum_{i \in I_{\text{nn}}} \hat{w}_i\frac{\epsilon_i + \epsilon_{j^*(i)}}{2}\right| + \sum_{i \in I_{\text{nn}}} |\hat{w}_i| a\|X_{j^*(i)} - (-X_i)\|^\alpha.\end{align*}

To address the fact that $\mu_e(x^*)$ is unobservable, we bound it using the closest observation $j^* = \arg\min_{j} \|X_j - x^*\|$. By H\"older continuity: $$|\mu_e(x^*)| \leq |\mu_e(X_{j^*})| + a\|X_{j^*} - x^*\|^\alpha.$$ For $j^* \in I_{\text{paired}}$, $$|\mu_e(X_{j^*})| \leq \left|\frac{Y_{j^*} + Y_{-j^*}}{2}\right| + \left|\frac{\epsilon_{j^*} + \epsilon_{-j^*}}{2}\right|.$$ A similar analysis holds for $j^* \in I_{\text{nn}}$.

Finally for the noise terms, we apply concentration with $\delta/4$ allocation: $$\left|\sum_{i=1}^n \hat{w}_i \epsilon_i\right| \leq \sigma\|\hat{w}\|_2\sqrt{2\log(6/\delta)},$$ $$\left|\sum_{i \in I_{\text{paired/nn}}} \hat{w}_i\frac{\epsilon_i + \epsilon_{j}}{2}\right| \leq \frac{\sigma}{\sqrt{2}}\left\|\hat{w}_{I_{\text{paired/nn}}}\right\|_2\sqrt{2\log(6/\delta)}$$ since $\frac{\epsilon_i + \epsilon_j}{2}$ is sub-Gaussian with parameter $\frac{\sigma}{\sqrt{2}}$, $$\left|\frac{\epsilon_{j^*} + \epsilon_{-j^*}}{2}\right| \leq \sigma\sqrt{2\log(6/\delta)}.$$ By union bound, all hold with probability $1-\delta$. The final result follows by combining each constituent term.
\end{proof}

\subsection{Proof of proposition \ref{prop:regularized_bound}}
\begin{proof}
By Proposition~\ref{prop: err}, with probability at least $1 - \delta$:
\begin{multline}
|\mu(x^*) - \hat{\mu}(x^*)| \leq \left|\mu(x^*) - \sum_{i=1}^n |\hat{w}_i|\,\mu(\X_i^\ddagger)\right| \\
+ 2\sum_{i=1}^n |\hat{w}_i|\,\mathbf{1}(\hat{w}_i < 0)\,a\|\X_i\|^\alpha + \sigma\|\hat{w}\|_2\sqrt{2\log(2/\delta)}
\end{multline}
The improvement in the noise bound follows from bounding $\|\hat{w}\|_2$ via the first-order optimality conditions. For each $i$:
\begin{equation}
2\sum_{j=1}^n \hat{w}_j X_j^T X_i - 2(x^*)^T X_i + 2\lambda\hat{w}_i + \gamma \|X_i\|^{\alpha} \xi_i = 0
\end{equation}
where $\xi_i \in \partial (|w_i| \mathbf{1}(w_i < 0))|_{w_i = \hat{w}_i}$. Multiplying by $\hat{w}_i$ and summing over all $i$:
\begin{equation}
2\left\|\sum_{j=1}^n \hat{w}_j X_j\right\|_2^2 - 2(x^*)^T \sum_{i=1}^n \hat{w}_i X_i + 2\lambda \|\hat{w}\|_2^2 + \gamma \sum_{i=1}^n |\hat{w}_i| \mathbf{1}(\hat{w}_i < 0) \|X_i\|^{\alpha} = 0
\end{equation}
where the sign of the $\gamma$ term is positive because for $\hat{w}_i < 0$: $\xi_i = -1$ (subgradient of $\max(0,-w_i)$), so $\hat{w}_i \cdot \gamma\|X_i\|^\alpha \cdot \xi_i = (-|\hat{w}_i|)\cdot\gamma\|X_i\|^\alpha\cdot(-1) = +|\hat{w}_i|\gamma\|X_i\|^\alpha$.

Rearranging: $2\lambda\|\hat{w}\|_2^2 = 2(x^*)^T\sum_i\hat{w}_iX_i - 2\bigl\|\sum_j\hat{w}_jX_j\bigr\|_2^2 - \gamma\sum_i|\hat{w}_i|\mathbf{1}(\hat{w}_i<0)\|X_i\|^\alpha$. Applying Young's inequality $(x^*)^T \sum_i \hat{w}_i X_i \leq \frac{1}{2}\|x^*\|_2^2 + \frac{1}{2}\|\sum_i \hat{w}_i X_i\|_2^2$:
\begin{align*}
2\lambda\|\hat{w}\|_2^2 &\leq \|x^*\|_2^2 + \bigl\|\textstyle\sum_i\hat{w}_iX_i\bigr\|_2^2 - 2\bigl\|\textstyle\sum_j\hat{w}_jX_j\bigr\|_2^2 - \gamma\textstyle\sum_i|\hat{w}_i|\mathbf{1}(\hat{w}_i<0)\|X_i\|^\alpha \\
&= \|x^*\|_2^2 - \bigl\|\textstyle\sum_i\hat{w}_iX_i\bigr\|_2^2 - \gamma\textstyle\sum_i|\hat{w}_i|\mathbf{1}(\hat{w}_i<0)\|X_i\|^\alpha \;\leq\; \|x^*\|_2^2
\end{align*}
since both subtracted terms are non-negative. Therefore:
\begin{equation}
\|\hat{w}\|_2 \leq \frac{\|x^*\|_2}{\sqrt{2\lambda}}
\end{equation}
This bound notably does not depend on $\gamma$ or the support radius $R$: increasing $\gamma$ further constrains the feasible set and cannot inflate $\|\hat{w}\|_2$. Substituting into Proposition~\ref{prop: err} completes the proof.
\end{proof}

\section{SYNTHETIC DATA STUDY}
\label{sec: simulation}
\subsection{Challenge Study}
In this section, we study the performance of our estimator via two data generative process (DGP), one using a linear DGP and the second one using a non-linear DGP. In this simulation study, we specifically consider a challenging case when the target point $\x^\star$ is outside the convex hull of the training points $\{\X_1,\dots,\X_n\}$. For linear DGP, the outcome $Y$ is a linear function of $\X$ while for nonlinear DGP, the outcome is a quadratic function of $\X$. The two DGPs are as follows:
\begin{align*}
    \text{Linear DGP:} & \mu(\X) = \beta^T \X \\
    \text{Nonlinear DGP:} & \mu(\X) = 2X_1^2 + X_2 + X_1 X_2 + \epsilon
\end{align*}
To further make the setup more challenging, we consider a relatively small sample with 10 training units and a target unit as shown in Figure~\ref{fig: convex_hull} (in main text). This scenario is particularly interesting because of the limited sample size compared to the problem's dimensionality: $n/p = 5$.

\begin{figure}
    \centering
    \begin{tabular}{cc}
    \includegraphics[width=0.49\linewidth]{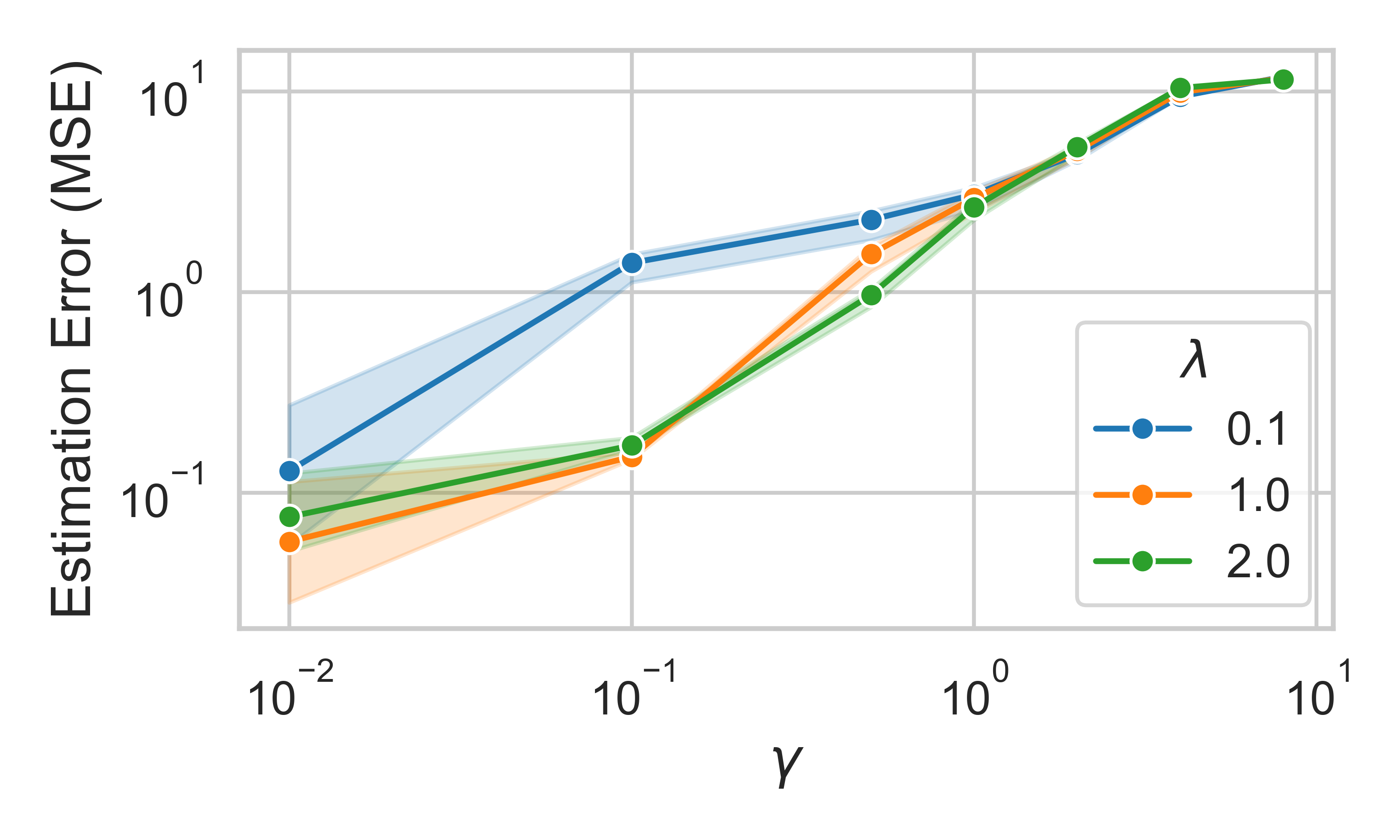} &
    \includegraphics[width=0.49\linewidth]{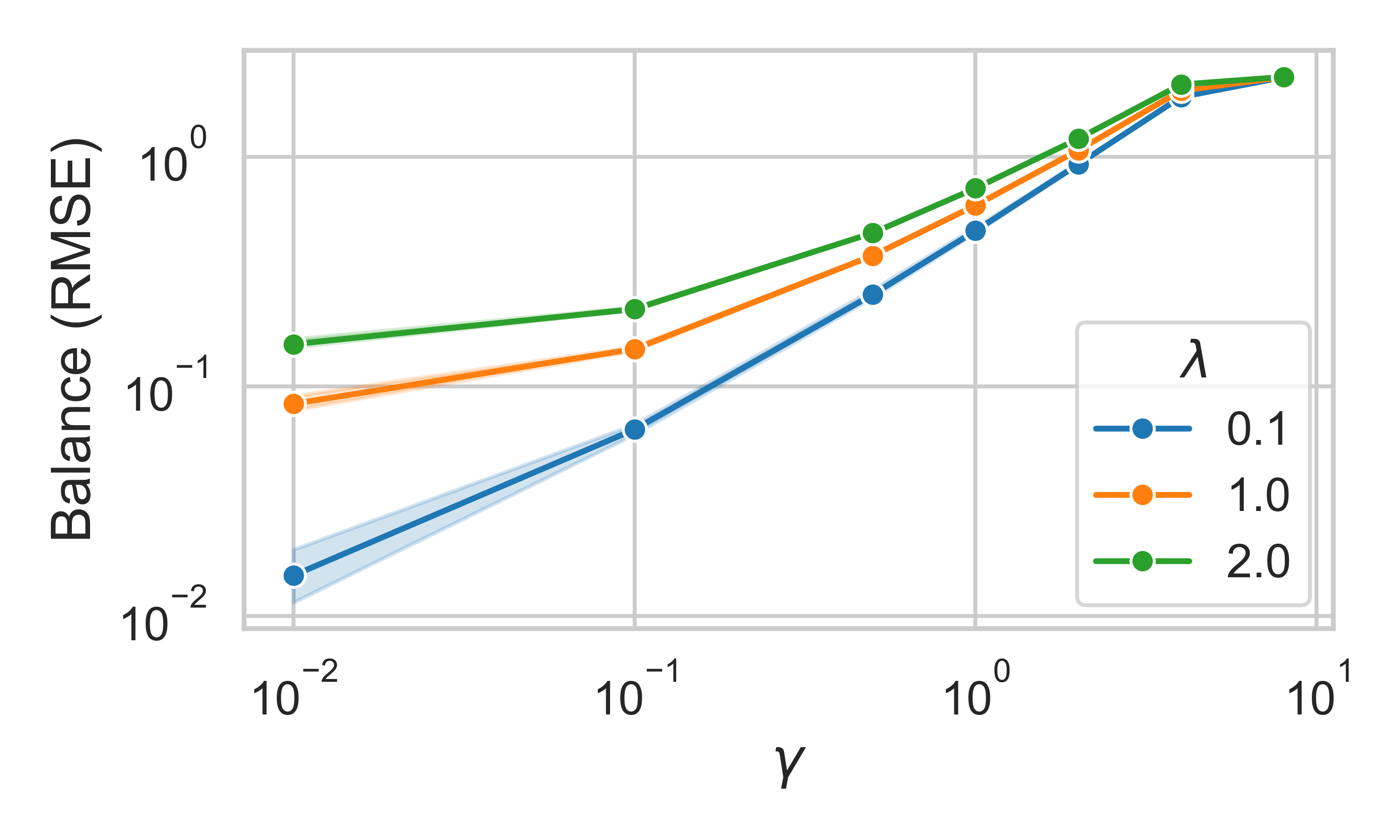}\\
    (a) & (b) \\
    \includegraphics[width=0.49\linewidth]{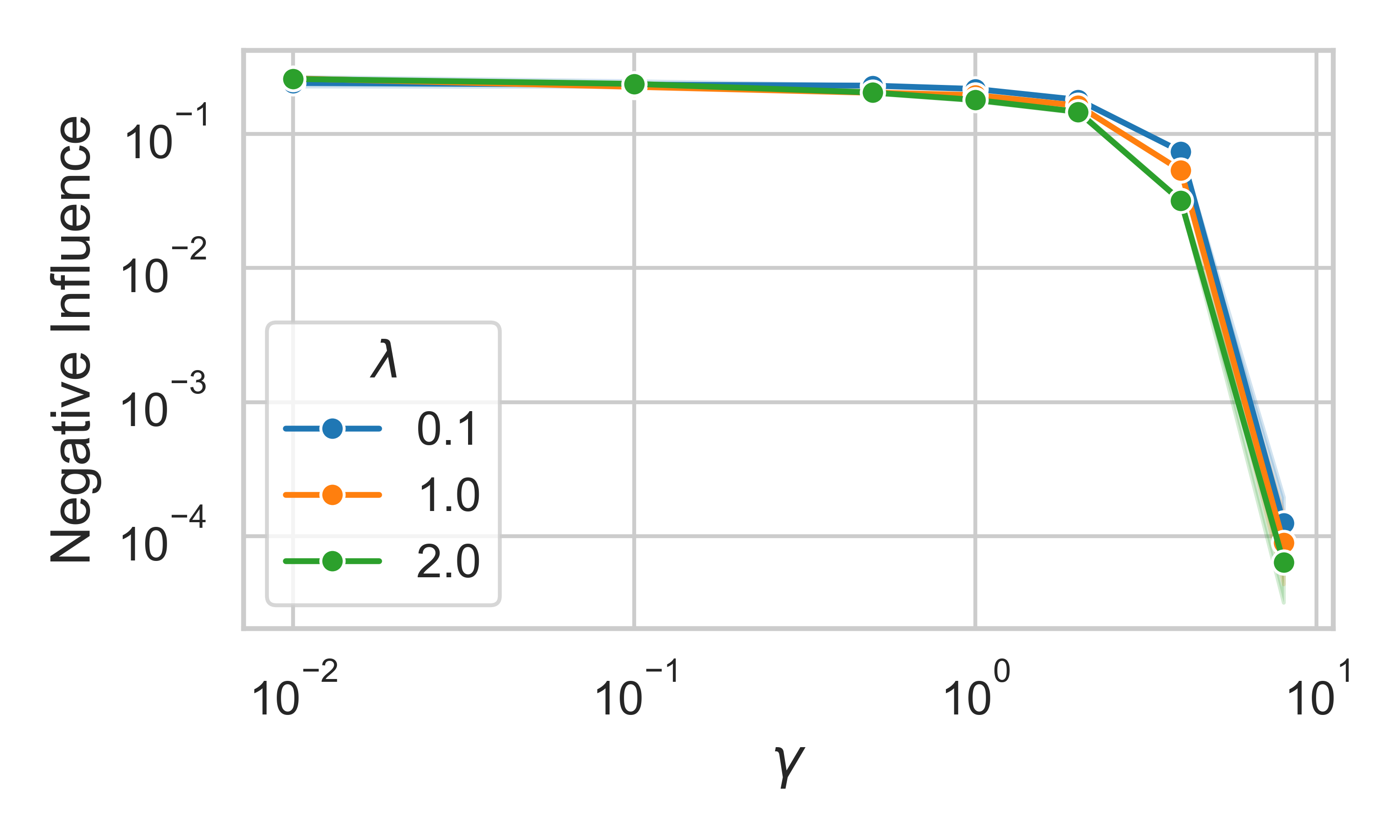} &
    \includegraphics[width=0.49\linewidth]{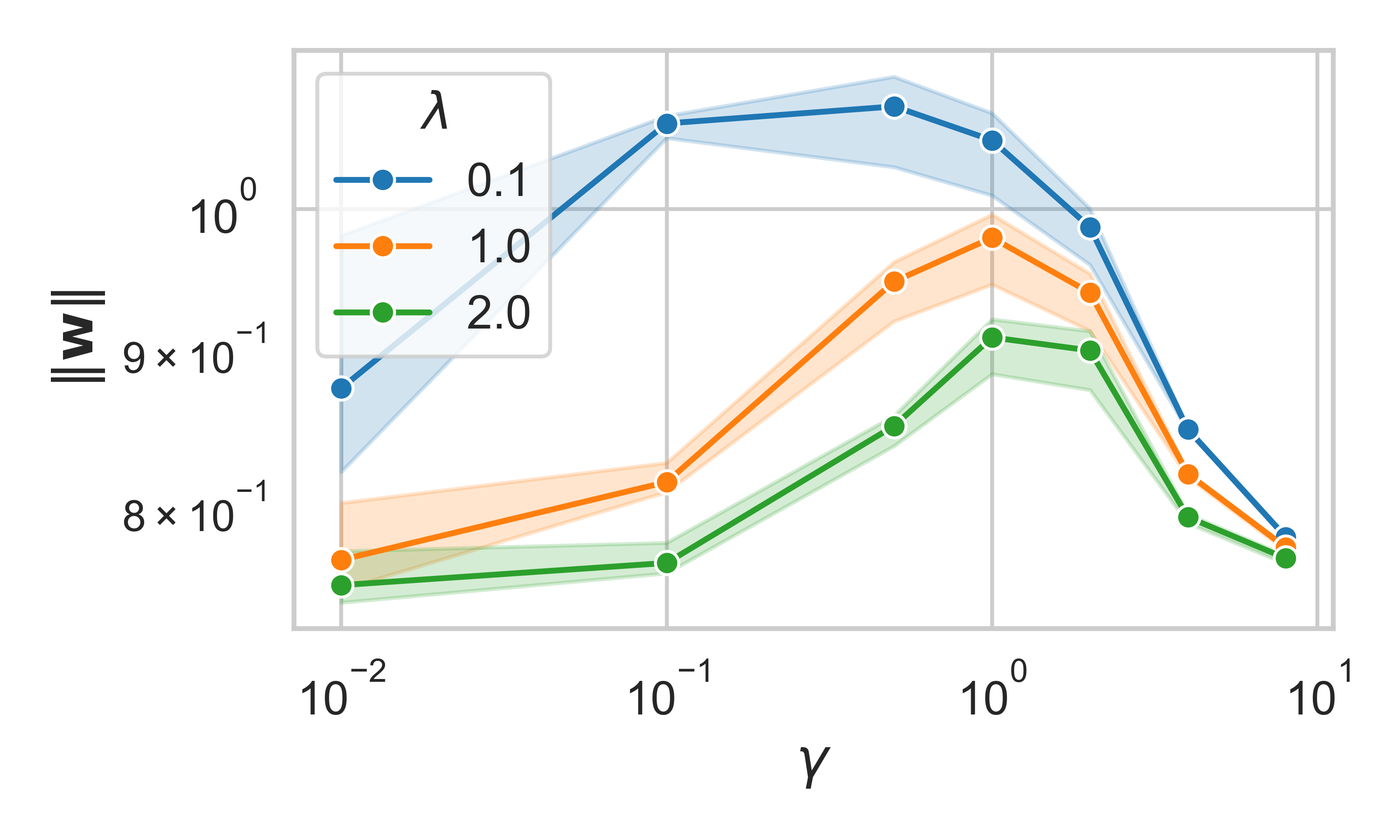} \\
    (c) & (d)
    \end{tabular}
    \caption{Results on synthetic data generated using linear DGP. (a) Estimation error measured as mean squared error, (b) balance between the weighted source and target populations, (c) extent of extrapolation measured as negative influence -- contribution on units with negative weights, (d) L2 norm $\mathbf{w}$ capturing asymptotic variance. }
    \label{fig:sim_linear}
\end{figure}

For the linear DGP, our parametric assumption holds. We observe that increasing the regularization on extrapolation (\(\gamma\)) decreases the negative influence while increasing the balance RMSE, as shown in Figures~\ref{fig:sim_linear}(b) and (c) -- consistent with theoretical expectations. As underlying DGP is linear, relying on parametric assumptions for extrapolation yields optimal estimates with the smallest estimation error corresponding to least level of regularization on extrapolation (see Figure~\ref{fig:sim_linear}(a)).

\begin{figure}
    \centering
    \begin{tabular}{cc}
    \includegraphics[width=0.49\linewidth]{figures/sim_res_nl_Error.png} &
    \includegraphics[width=0.49\linewidth]{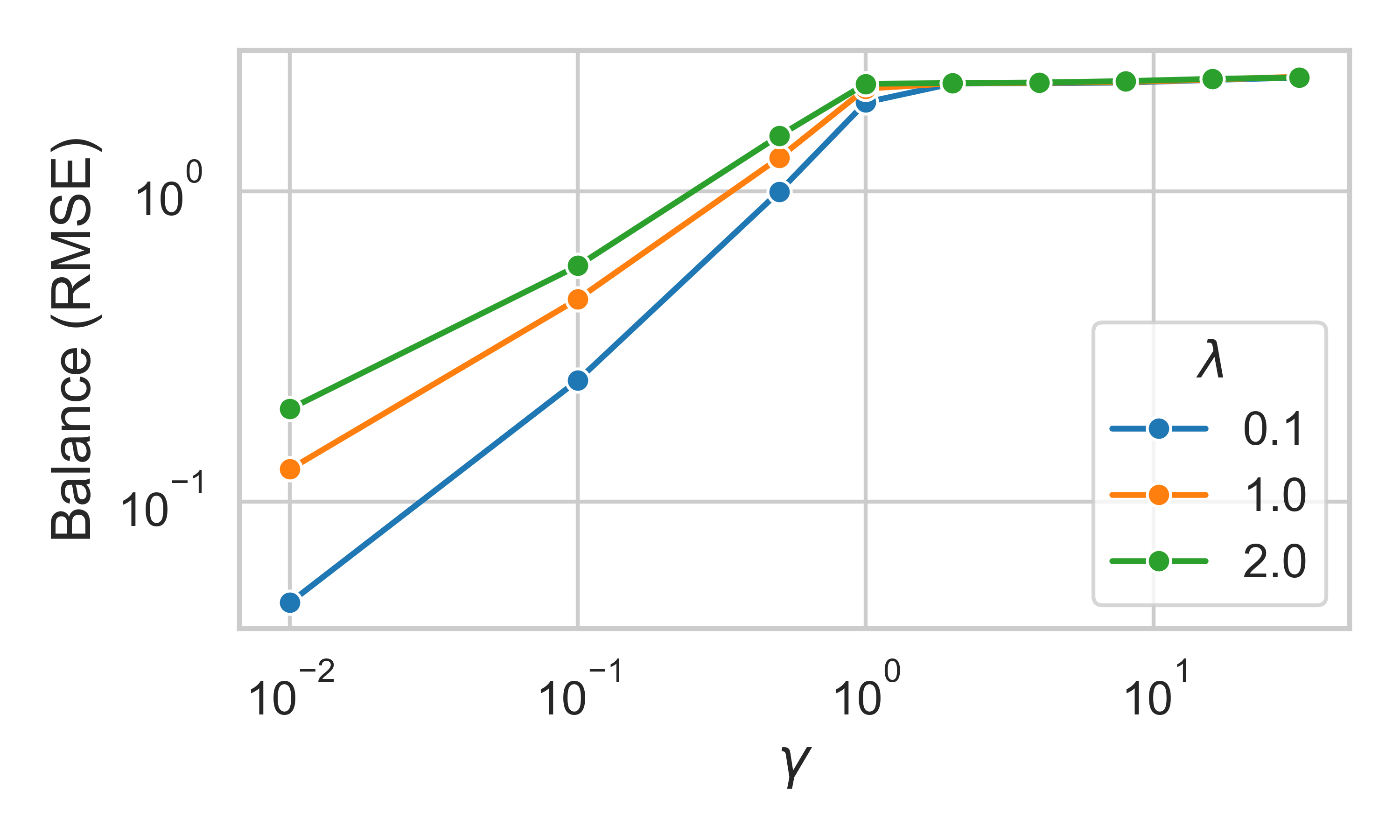}\\
    (a) & (b) \\
    \includegraphics[width=0.49\linewidth]{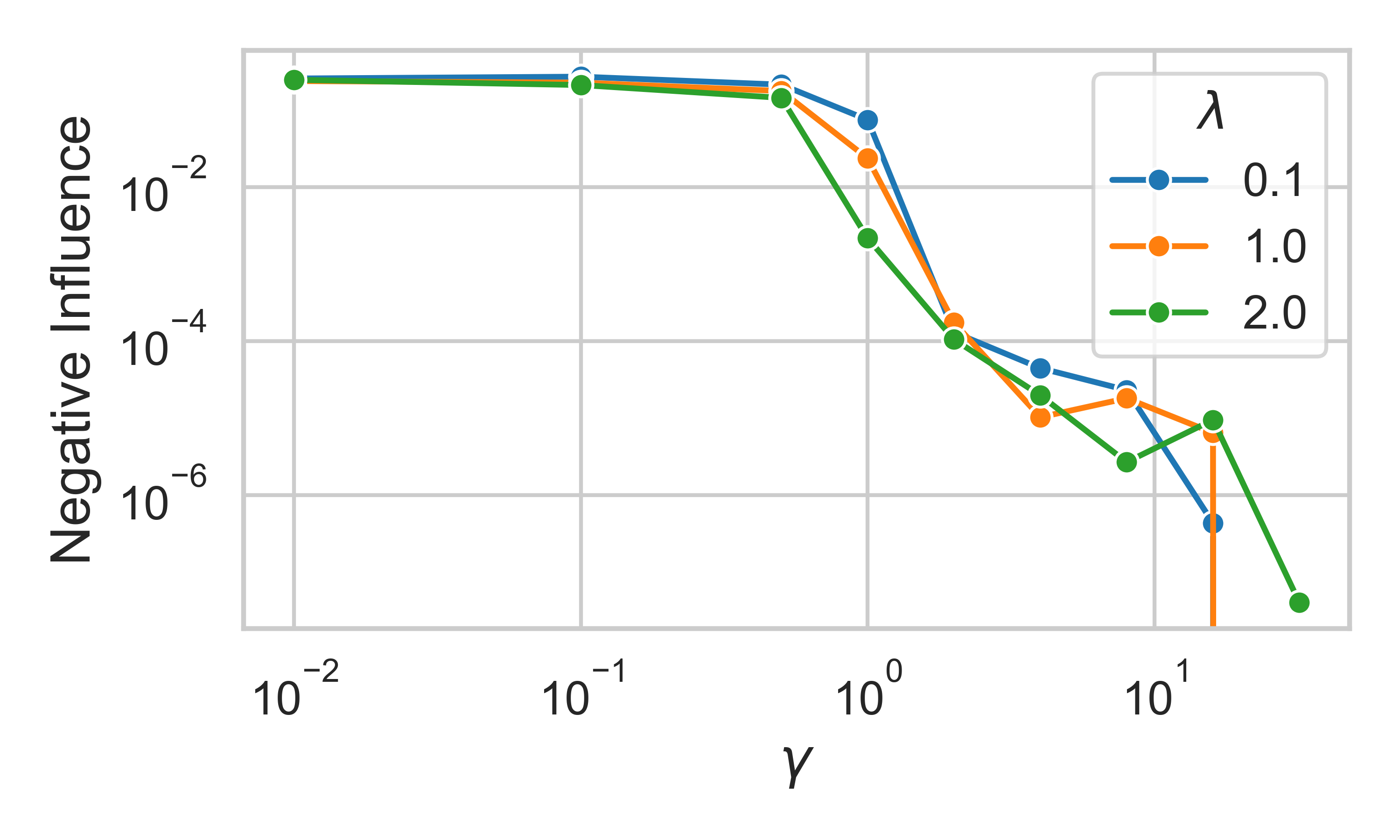} &
    \includegraphics[width=0.49\linewidth]{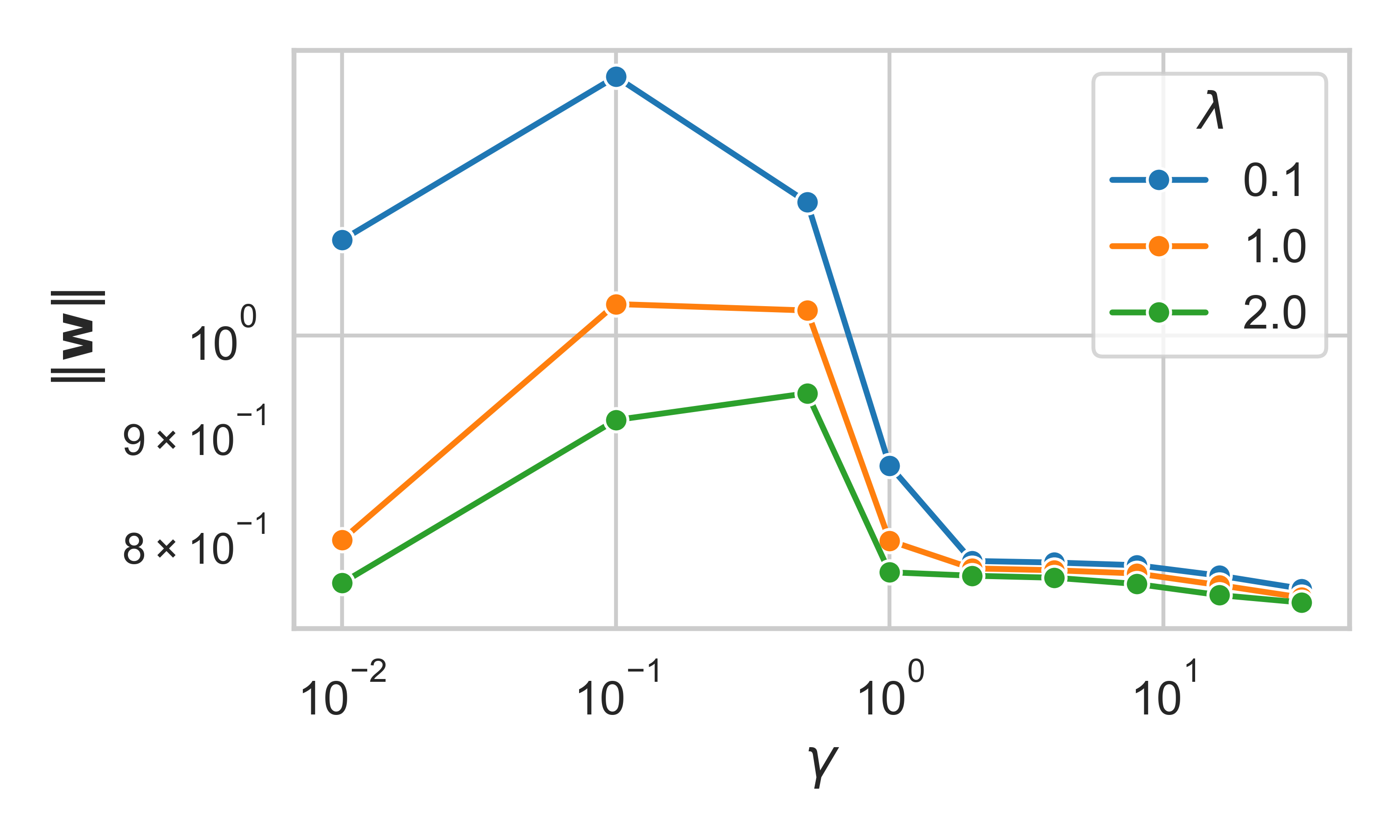} \\
    (c) & (d)
    \end{tabular}
    \caption{Results on synthetic data generated using non-linear DGP. (a) Estimation error measured as mean squared error, (b) balance between the weighted source and target populations, (c) extent of extrapolation measured as negative influence -- contribution on units with negative weights, (d) L2 norm $\mathbf{w}$ capturing asymptotic variance. }
    \label{fig:sim_nonlinear}
\end{figure}

Unlike linear DGP, the outcome function in the nonlinear DGP is not an odd function and thus the parametric assumption is violated. The outcome function has a quadratic term, an interaction term, and a linear term. Intuitively, we expect that a small amount of extrapolation might be beneficial due to the linear component however large amount of extrapolation may result in a high error rate due to violation of parametric assumption. The estimation error rate shown in Figure~\ref{fig:sim_nonlinear}(a) is in congruency with the above-discussed expectation -- thus highlighting the tradeoff between two different kinds of biases.

\subsection{High-dimensional Large Scale Study}
\begin{figure}
    \centering
    \includegraphics[width=0.7\linewidth]{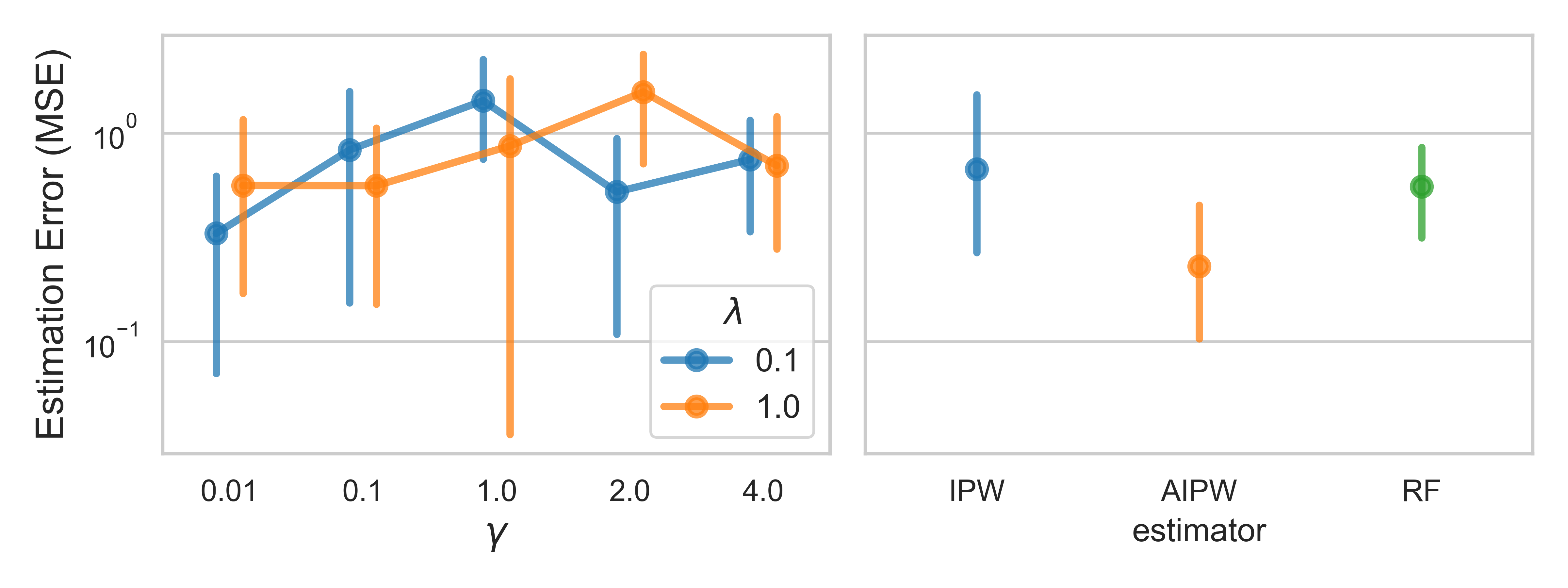}
    \caption{Results on synthetic data generated using High-dimensional Friedman's DGP.}
    \label{fig:friedman}
\end{figure}

We now expand our analysis to a large scale high-dimensional data. In particular, we consider a scenario where $p/n$ is large. We generate synthetic data via Friedman DGP which is commonly used across machine learning and causal inference literature. We generate $n_{control}=1000$ control units and $n_{treat}=1000$ with $p=5000$ covariates. The DGP is Friedman DGP is defined as follows:
$$Y(0) = 10 * \sin(\pi X_0 X_1) + 20  (X_2 - 0.5) ^ 2 + 10  X_3 + 5  X_4 + \epsilon,$$
and $Y(1) = Y(0)$. In this example, we are interested in estimating average treatment effect on the treated -- the truth is $0$. Here, the control units are used as the training set estimate counterfactual for the treated units which act as estimation set here -- $n/p = 1000/5000 = 0.2$. We compare our approach with IPW, AIPW and Random Forest (RF). One main difference between our framework and AIPW or RF is that AIPW and RF both uses outcome information to learn the estimator while our framework (similar to IPW) is outcome data agnostic and only uses covariate information to ensure balance while also regularizing extrapolation. Figure~\ref{fig:friedman} shows the mean squared error in estimating ATT and compares it with IPW, AIPW and RF. Our results show that sweeping over the range of $\gamma$ (i.e., regularizing extrapolation) smoothly `interpolates' between AIPW and IPW estimators (the former allows extrapolation using a linear regression outcome model, while the latter relies only on interpolation).

\end{document}